%% file: main_SDW.tex
\documentclass[10pt]{article} 
\usepackage[preprint]{tmlr}

\input{math_commands.tex}

\usepackage{algorithm}
\usepackage{algpseudocode}

\usepackage{hyperref}
\usepackage{url}
\usepackage{graphicx}
\usepackage{booktabs}

\usepackage{tabularx}
\usepackage{array}

\usepackage{amsthm}
\newtheorem{proposition}{Proposition}

\usepackage{placeins}

\title{Stochastic Diversity Windows: Diagnosing Noise Influence in Robotic Diffusion Sampling}

\author{
\name
Tina Lakhani\textsuperscript{1} \quad
Jing Wu\textsuperscript{1} \quad
Yu-Kun Lai\textsuperscript{1} \quad
Ze Ji\textsuperscript{2} \\
\addr
\textsuperscript{1}School of Computer Science and Informatics,
Cardiff University, UK \\
\textsuperscript{2}School of Engineering,
Cardiff University, UK \\
\email lakhaniif@cardiff.ac.uk
}

\def\month{MM}  
\def\year{YYYY} 
\def\openreview{\url{https://openreview.net/forum?id=XXXX}} 

\begin{document}

\maketitle
\begin{abstract}
Diffusion models are increasingly adopted in robotics and control due to their capability to express multiple feasible solutions to a given task. However, the role of stochasticity in the diffusion model--sampler across denoising remains poorly understood. We introduce a per-timestep counterfactual diagnostic that tests how changing the sampled noise affects the final output. Across five model--sampler configurations spanning three robotics benchmarks, diverse architectures, domains, and conditioning mechanisms, we reveal highly varied temporal sensitivity profiles, where noise-induced deviations can amplify, partially contract, or undergo near-total attenuation downstream. From these profiles, we define \textit{stochastic diversity windows} (SDWs): compact intervals that identify where sampler stochasticity has the strongest measured influence on the final output. Guided by these windows, we apply two heuristic interventions: targeted stochastic amplification, which increases sample diversity exploration of the solution space, and an \textit{SDW-hybrid} sampler, which replaces DDPM steps with DDIM skips in low-impact regions, resulting in network evaluation reductions of up to 40--60\% and wall-clock time by up to 85\%, with no consistent changes detected in feasibility, cost, or diversity across held-out tasks.
\end{abstract}

\section{Introduction}
Many robotic applications, from motion planning to visuomotor control, benefit from multiple valid solutions under the same task constraints. By sampling directly from learned conditional distributions, diffusion models~\citep{ho2020denoising,nichol2021improved,song2021scorebased} can synthesize rich families of feasible trajectories conditioned on start and goal states, obstacle geometries, or visual observations~\citep{pmlr-v162-janner22a,zhu2023diffusion,Wolf2025Diffusion,shao2025generative,Yin2025Diverse,Rosasco2025KDPE}.
Stochastic perturbations during sampling can promote diverse trajectory generation by altering the evolving sample or the information presented to subsequent model steps~\citep{karras2022elucidating,sadat2024cads}.

Because these updates follow a learned score field~\citep{alain2014regularized,vincent2011connection,song2021scorebased}, the downstream effect of an individual stochastic perturbation cannot be assumed to be preserved through subsequent denoising. Our diagnostic counterfactually probes each denoising timestep by resampling its injected noise and measuring the resulting clean output change. Across configurations, this reveals different temporal profiles in which stochastic influence may be strongly attenuated, persist, or increase downstream.

We summarize each profile with \emph{stochastic diversity windows} (SDWs): the shortest contiguous timestep ranges that capture a chosen fraction of the measured influence. Each window is estimated across multiple diagnostic tasks to obtain a configuration-level intervention region. We evaluate the diagnostic across five model--sampler--conditioning configurations spanning three robotic benchmarks: Maze2D~\citep{pmlr-v162-janner22a}, Motion Planning with Diffusion (MPD)~\citep{carvalho2025motion}, and Diffusion Policy for PushT~\citep{Chi2024Diffusion}. The resulting windows are then held fixed and used to guide inference-time interventions across held-out tasks, without recalculating the diagnostic for each new task. 

In summary, our main contributions are:
\begin{itemize}
    \item We introduce a per-timestep counterfactual diagnostic for measuring the downstream endpoint influence of native sampler stochasticity. The resulting temporal sensitivity profiles are summarized by compact intervals (SDWs), capturing a specified fraction of the measured influence.

    \item Across five model--sampler--conditioning configurations spanning three robotic benchmarks, we show that the temporal survival of stochastic influence is strongly configuration dependent, with substantial differences in where perturbations remain consequential after subsequent denoising.

    \item We demonstrate that the diagnosed regions are actionable by using them to guide inference-time intervention. Targeted stochastic amplification produces its strongest diversity effects when focused on diagnosed high-influence regions, while SDW-guided stochastic--deterministic sampling removes computation preferentially from low-influence regions while preserving stochastic transitions where their measured downstream influence is greatest.
\end{itemize}

\section{Related Work}
\label{subsec:related_work}

Diagnosing and interpreting the denoising dynamics of diffusion models requires examining timestep-level behavior and the relationships between intermediate states and final outputs~\citep{lakhani2025methodologies}. Previous interpretability work has examined the emergence of perceptual and semantic representations across denoising~\citep{choi2022perception,tinaz2025emergence}, as well as whether concepts introduced at intermediate stages persist to the final output~\citep{gorgunTemporalConceptDynamics2026}. A related line of work considers how errors and perturbations propagate through denoising,~\citet{li2024error} study the propagation of denoiser approximation errors across successive updates, while \citet{strasman2026forgetting} establish stability conditions under which the influence of initialization mismatch, score-approximation error, and discretization errors are geometrically attenuated along the reverse process. From a step-wise perspective, \citet{pmlr-v306-li26ix} study contraction of discretized probability-flow ODE updates as a sufficient condition for perturbation suppression. Collectively, these studies demonstrate that the downstream consequence of an intermediate perturbation cannot be inferred from its immediate magnitude alone, since subsequent denoising may attenuate, or reshape its influence. SDW directly identifies and quantifies such characteristics by measuring where a stochastic sampler's native per-step noise injections retain consequential influence on the final output in a deployed model--sampler configuration.

Guidance methods for sampling control often work by scheduling, restricting, or reshaping conditional signals across denoising~\citep{NEURIPS2021_49ad23d1,ho2022classifier,dinh2023rethinking,kynkaanniemi2024applying,yehezkel2025navigating,jinStagewiseDynamicsClassifierFree2026,chung2024cfg++,NEURIPS2025_3ec6c6fc}. Other approaches alter conditioning noise, initial latents, score temperature, or additional sampling-time perturbations~\citep{sadat2024cads,manneringNoiseLevelDiffusionGuidance2025,xu2025temporal,wuTimeAnnealedPerturbationSampling2026}. \citet{hou2025omegance}, for example, vary the strength of the model-predicted denoising term across timesteps, and \citet{karras2022elucidating} restrict Langevin-like ``churn'' to selected noise ranges, with the amount and location of additional stochasticity tuned empirically. Related work on efficient sampling similarly redistributes computation across the trajectory: DDIM introduces deterministic non-Markovian sampling~\citep{song2020denoising}, higher-order and timestep-selection methods reduce the number of denoiser evaluations~\citep{lu2022dpm,shen2025efficient,tai2025efficient,xu2025pretrained}, Restart Sampling alternates forward noise injection with deterministic integration~\citep{xu2023restart}, and \citet{liu2026hybriddps} combine DDPM and DDIM regimes using a fixed switch-point. These methods demonstrate that where stochasticity, guidance, or computation is placed can materially affect sampling behaviour, but the intervention schedule is generally prescribed, optimized, or tuned rather than derived from a direct measurement of where native stochasticity survives subsequent denoising.

Our focus is the stochastic sampler's own native noise and how its influence survives across denoising in a given configuration. By counterfactually resampling individual stochastic transitions and measuring their downstream final effect, SDW reveals a configuration-specific temporal sensitivity profile inherent to the sampler, without prior behavioral assumptions.

\section{Counterfactual Perturbation Diagnostic}
\label{sec:counterfactual_diagnostic}

Let \(\mathbf c\) denote the conditioning context of a single diffusion sampling call. For Maze2D and MPD, \(\mathbf c\) is fixed by the task start--goal condition; for PushT, \(\mathbf c_j\) denotes the observation--conditioning context at replan \(j\) within a closed-loop episode. The sampler adds a perturbation \(\boldsymbol{\eta}_t=\sigma_t\mathbf z_t\), where \(\mathbf z_t\sim\mathcal N(\mathbf 0,\mathbf I)\) is a Gaussian random variable and \(\sigma_t\) determines the perturbation magnitude. A diffusion sampler \(S_\theta\) maps the conditioning context, an initial latent \(\mathbf x_T\sim\mathcal N(\mathbf 0,\mathbf I)\), and the sequence \(\mathbf z_{1:T}=(\mathbf z_1,\dots,\mathbf z_T)\) to a final sample \(\mathbf x_0\):
\begin{equation}
\mathbf{x}_0 = S_\theta(\mathbf c,\mathbf x_T,\mathbf z_{1:T}).
\label{eq:sampler_mapping}
\end{equation}

For PushT, this mapping is applied separately at each replan using the corresponding context \(\mathbf c_j\). Depending on the application, \(\mathbf x_0\) may represent a trajectory, motion plan or action sequence. We use \(S_\theta\) to denote the complete diffusion sampling procedure for one model invocation, including the trained model, sampling algorithm, and any sampling-time conditioning or guidance, such as trajectory inpainting, classifier-free guidance~\citep{ho2022classifier}, or cost-guided planning~\citep{pmlr-v162-janner22a}.

\paragraph{Generating the Counterfactual.}
For each task \(\mathbf c\), we construct \(M\) baseline sampling calls indexed by \(m\). For Maze2D and MPD, these correspond to different stochastic outputs of the same fixed start--goal task. For PushT, they correspond to selected replans within a closed-loop episode; at each selected replan, the current observation history is fixed and a local baseline is constructed for that policy call. We probe the first ten re-planning contexts within each of ten closed-loop episodes.

For each baseline call and probed timestep \(t\), we replace only \(\mathbf z_t\) with an independent sample \(\mathbf z'_t\sim\mathcal N(\mathbf 0,\mathbf I)\), while holding that call's inputs, initial latent \(\mathbf x_T\), and all other noise draws fixed. We repeat this replacement \(R\) times, so each counterfactual differs from its baseline by exactly one isolated stochastic perturbation. Algorithm~\ref{alg:single_step_perturbation} summarizes the procedure.

\paragraph{Endpoint Sensitivity.}
For task \(\mathbf c\), let \(\mathbf x_0^{(\mathbf c,m)}\) denote baseline sample \(m\), and let \(\tilde{\mathbf x}_0^{(\mathbf c,m,t,r)}\) denote the corresponding counterfactual output produced by the \(r\)-th replacement at timestep \(t\). Because the baseline and counterfactual differ only in \(\mathbf z_t\), their final-output deviation measures how strongly that single-timestep perturbation influences the final clean sample.

We measure the resulting deviation between baseline and counterfactual at the end of denoising as
\[
\Delta\left( 
\mathbf x_0^{(\mathbf c,m)}, 
\tilde{\mathbf x}_0^{(\mathbf c,m,t,r)} 
\right) =
\left\| 
\phi\left(\mathbf x_0^{(\mathbf c,m)}\right) -
\phi\left(\tilde{\mathbf x}_0^{(\mathbf c,m,t,r)}\right)
\right\|_2,
\]

where \(\phi(\cdot)\) vectorizes each output.

For task \(\mathbf c\) and timestep \(t\), we estimate \textit{endpoint sensitivity} by averaging over the \(M\) baseline samples and \(R\) replacement draws:
\begin{equation}
\hat q_t(\mathbf c) =
\frac{1}{MR}
\sum_{m=1}^{M}
\sum_{r=1}^{R}
\Delta
\left(
\mathbf x_0^{(\mathbf c,m)},
\tilde{\mathbf x}_0^{(\mathbf c,m,t,r)}
\right).
\label{eq:timestep_sensitivity}
\end{equation}

For PushT, Equation~\ref{eq:timestep_sensitivity} therefore averages the \(R\) counterfactual replacements within each selected replan and then across replans, yielding one episode-level endpoint-sensitivity curve. Thus, \(\hat q_t(\mathbf c)\) estimates the expected final-output change caused by resampling the injected noise at a single timestep \(t\). 

We compute \(L_2\) distance in each benchmark's output representation, so the diagnostic measures the magnitude of the change in the generated trajectory, motion plan, or action sequence. 

\begin{figure}[t]

{\centering
\includegraphics[width=0.45\textwidth]{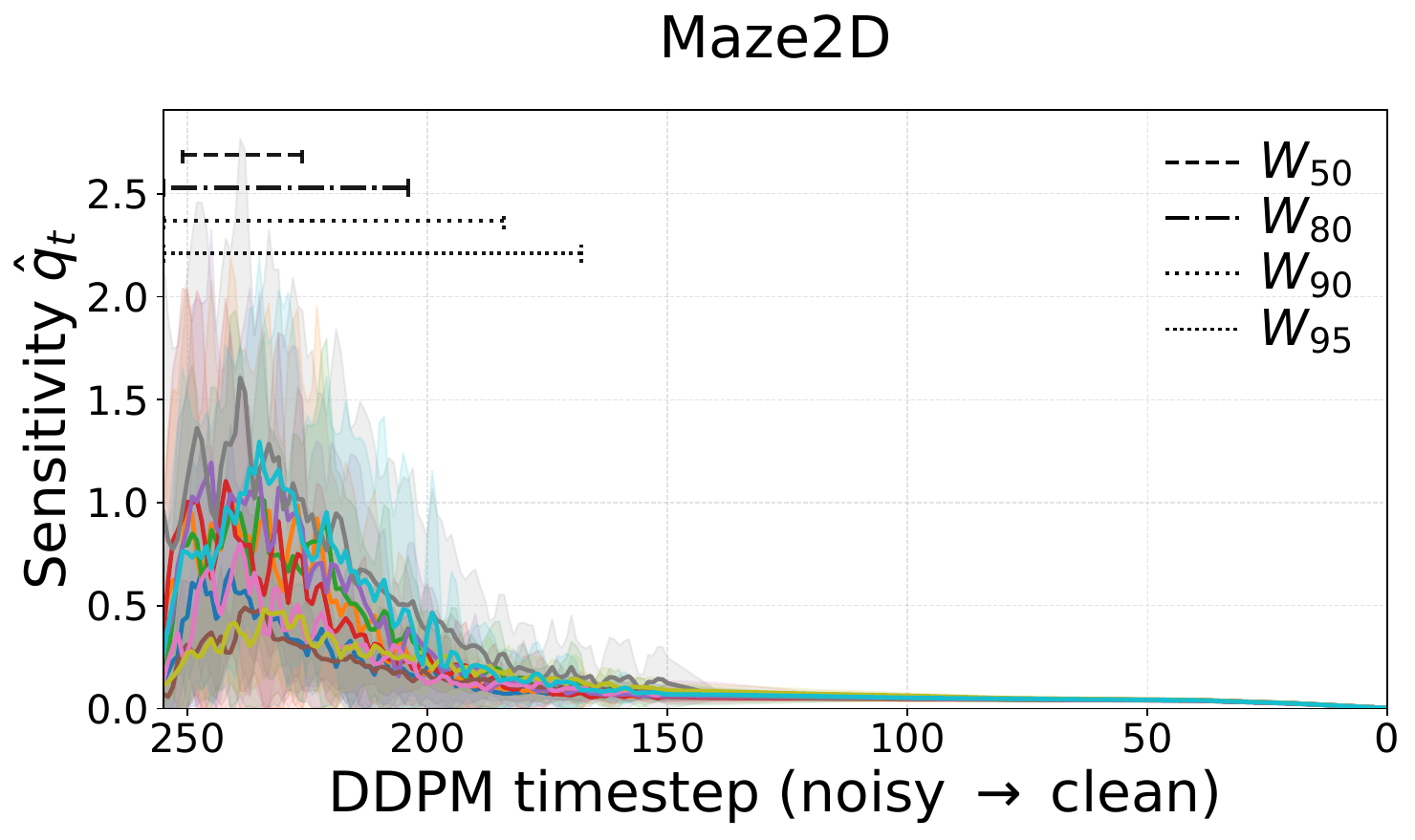}
\hfill
\includegraphics[width=0.45\textwidth]{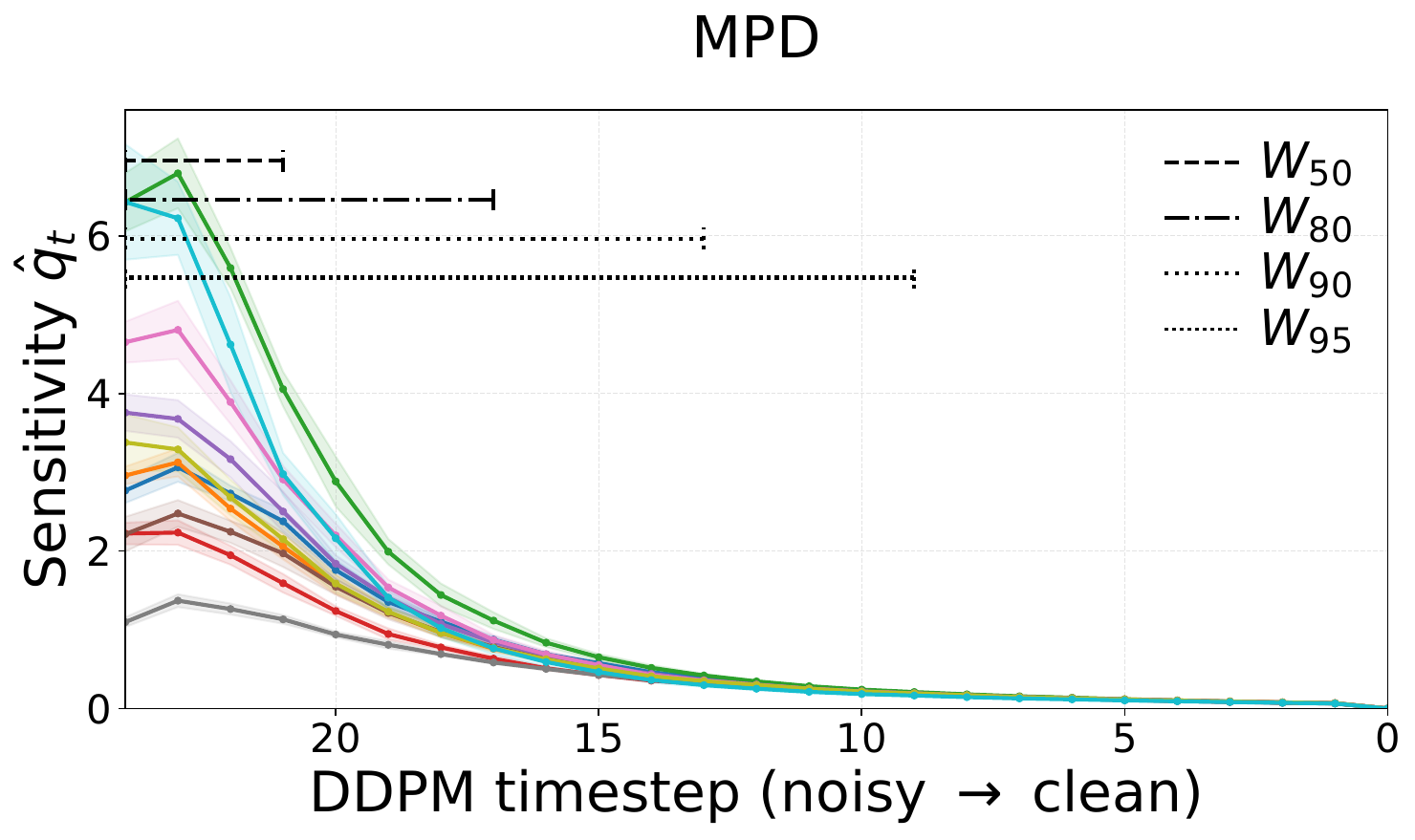}
\par
\includegraphics[width=0.45\textwidth]{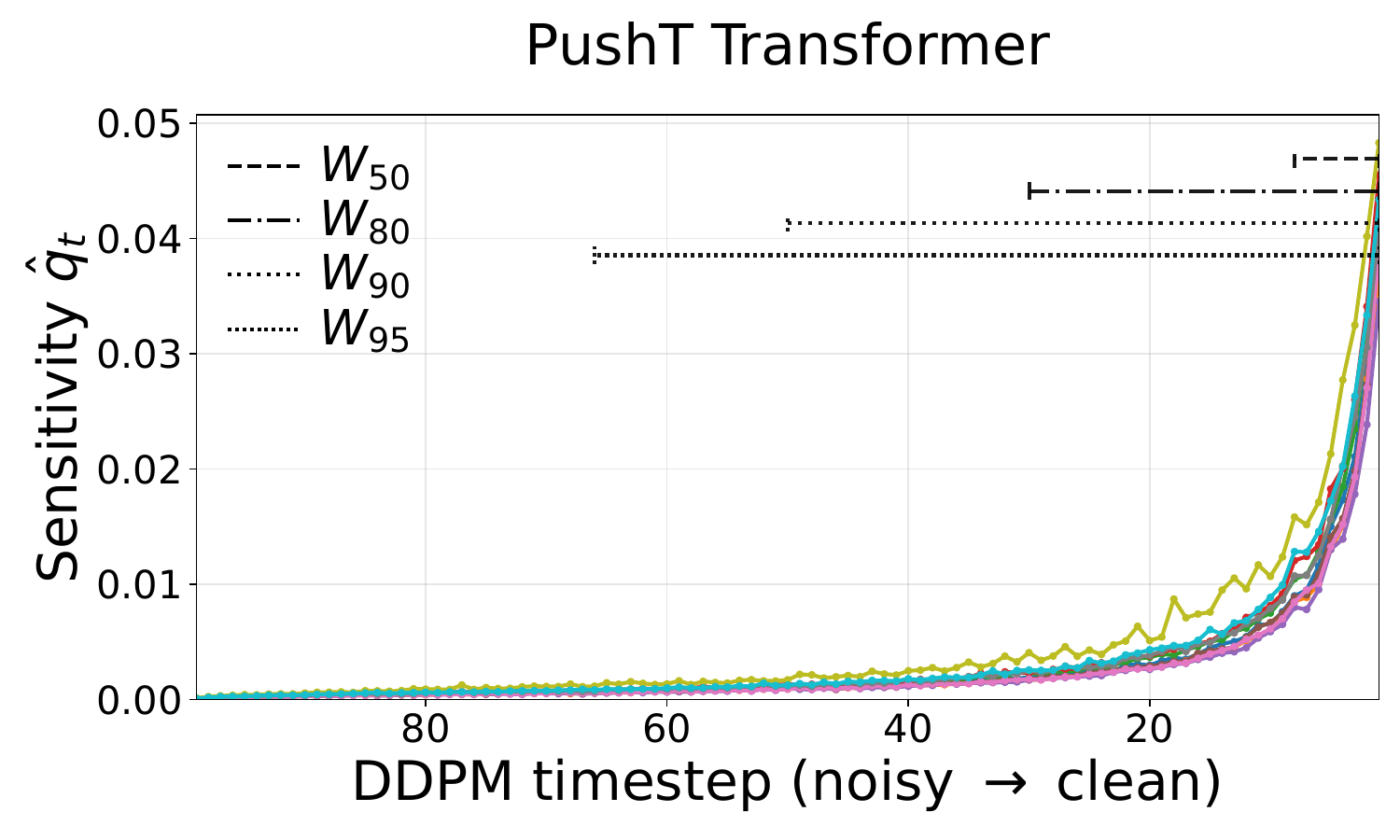}
\hfill
\includegraphics[width=0.45\textwidth]{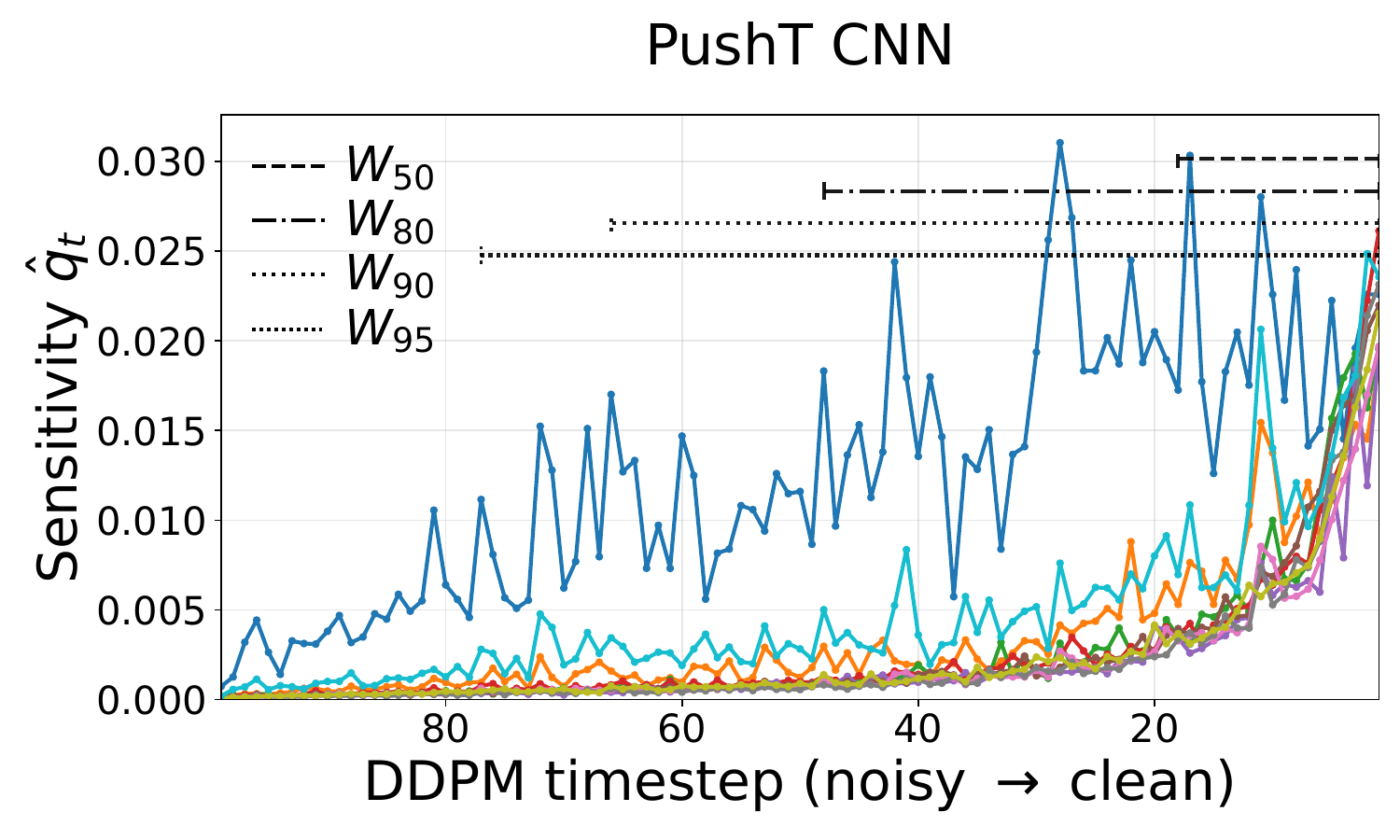}
\par
}

\noindent\textbf{(a)} Endpoint sensitivity and stochastic diversity windows\par

{\centering
\includegraphics[width=0.235\textwidth]{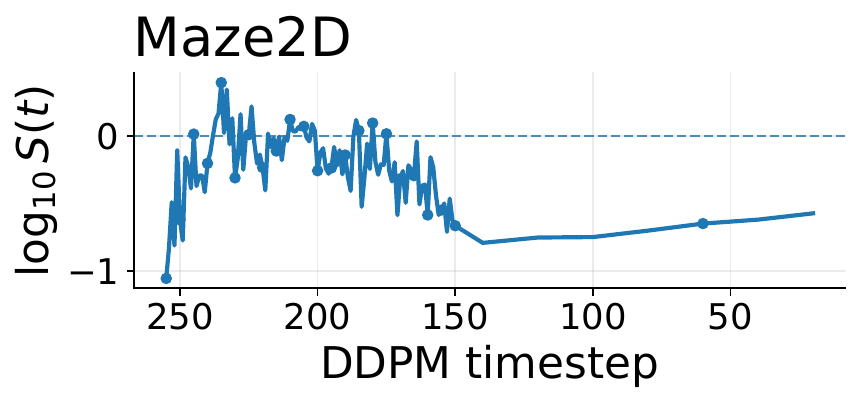}
\hfill
\includegraphics[width=0.235\textwidth]{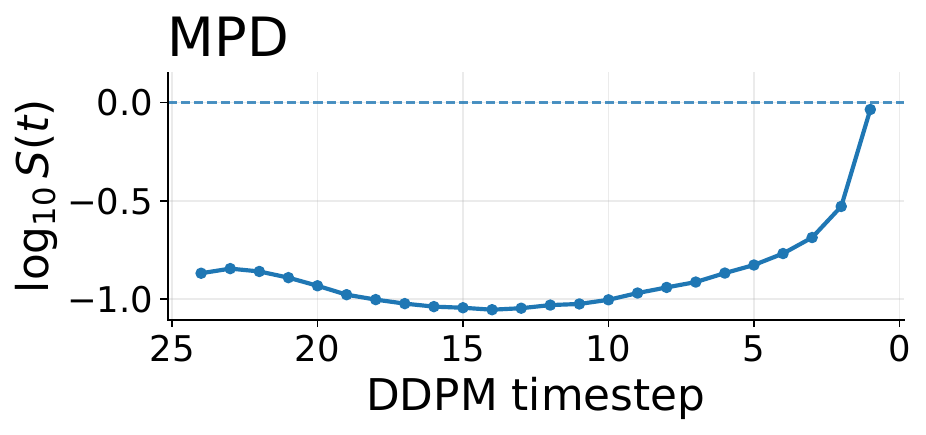}
\hfill
\includegraphics[width=0.235\textwidth]{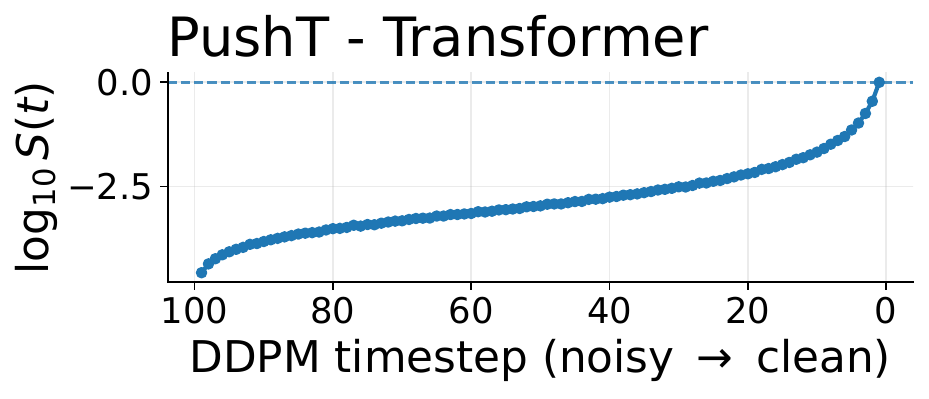}
\hfill
\includegraphics[width=0.235\textwidth]{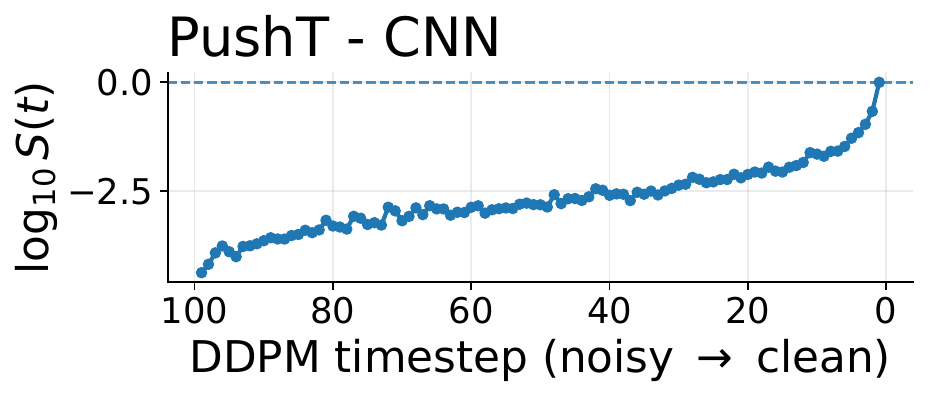}
\par
}

\noindent\textbf{(b)} Survival of first affected perturbations\par

\caption{Diagnostic curves. (a) Each subplot shows endpoint sensitivity \(\hat q_t\) for one model--sampler configuration. The ten coloured curves correspond to ten tasks, with each curve averaged over baseline seeds and replacement draws. The overlays mark stochastic diversity windows \(W_\alpha\) computed from the mean curve across those tasks. (b) Survival-ratio curves are plotted as \(\log_{10} S(t)\). The dashed line at \(\log_{10}S(t)=0\) corresponds to \(S(t)=1\). Values below the line indicate contraction of the perturbation-induced deviation, while values above indicate amplification.}
\label{fig:sensitivity-curves}
\end{figure}

We then average the task-level sensitivity curves over the whole diagnostic task set \(\mathcal C\):

\[
\bar q_t
=
\frac{1}{|\mathcal C|}
\sum_{\mathbf c\in\mathcal C}
\hat q_t(\mathbf c).
\]
We use this aggregate curve to construct the configuration-level stochastic diversity windows used in Figure~\ref{fig:sensitivity-curves} and in the subsequent interventions.

Proof that Equation~\ref{eq:timestep_sensitivity} is an unbiased Monte Carlo estimator is provided in Appendix~\ref{app:mc_unbiased}.

\paragraph{Stochastic Diversity Windows.}

The diagnostic yields a per-task endpoint-sensitivity curve \(\hat q_t(\mathbf c)\) over denoising. Because \(\hat q_t\) measures how stochastic perturbations at each timestep propagate to variation in the final output, we use it to identify denoising regions in which sampler stochasticity has the greatest downstream influence. We call these regions \textit{stochastic diversity windows}. Intervention outcomes are evaluated later using feasible-sample diversity \(D\) (Equation~\ref{eq:feasible-diversity}), which measures pairwise variation among feasible samples in a generated batch. We define \(W_\alpha\) as the shortest contiguous interval containing at least a fraction \(\alpha\in(0,1)\) of the total aggregate endpoint sensitivity:

\begin{equation}
\label{eq:capture-window}
W_\alpha
=
\arg\min_{W\in\mathcal W}|W|
\quad
\text{s.t.}\quad
\sum_{t\in W}\bar q_t
\ge
\alpha\sum_{t=1}^{T}\bar q_t,
\end{equation}
where \(\mathcal{W}\) is the set of all contiguous timestep intervals.
If more than one interval has the same minimum length, we select the interval of greatest total sensitivity. If a tie still remains, we choose the interval closer to the start of denoising. 

Thus, contiguity provides a compact summary of where stochastic sensitivity is concentrated, with the interval location determined directly by the measured sensitivity profile.

\begin{algorithm}[t]
\caption{Single-step counterfactual perturbation}
\label{alg:single_step_perturbation}
\begin{algorithmic}[1]
\Require task instances \(\mathcal{C}\), baseline seeds \(M\), perturbation repeats \(R\),
sampler \(\mathcal{S}_\theta\), deviation function \(\Delta\)

\For{\(\mathbf{c}\in\mathcal{C}\)}
    \For{\(m=1,\ldots,M\)}
        \State Sample \(\mathbf{X}_T^{(\mathbf c,m)}\)
        and \(\mathbf{Z}_{1:T}^{(\mathbf c,m)}\)

        \State Generate baseline \(\mathbf{X}_0^{(\mathbf c,m)}\)
        using Eq.~\ref{eq:sampler_mapping}

        \For{\(t=1,\ldots,T\)}
            \For{\(r=1,\ldots,R\)}
                \State Set
                \(\tilde{\mathbf{Z}}_{1:T}
                \leftarrow
                \mathbf{Z}_{1:T}^{(\mathbf c,m)}\)

                \State Resample
                \(\tilde{\mathbf{Z}}_t
                \sim\mathcal{N}(\mathbf{0},\mathbf{I})\)

                \State Generate
                \(\tilde{\mathbf{X}}_0^{(\mathbf c,m,t,r)}
                =
                \mathcal{S}_\theta(
                \mathbf c,
                \mathbf{X}_T^{(\mathbf c,m)},
                \tilde{\mathbf{Z}}_{1:T})\)

                \State Record
                \(\Delta(
                \mathbf{X}_0^{(\mathbf c,m)},
                \tilde{\mathbf{X}}_0^{(\mathbf c,m,t,r)})\)

                \State Record the first affected response
                \(\Delta_{\mathrm{first}}^{(\mathbf c,m,t,r)}\)
            \EndFor
        \EndFor
    \EndFor

    \State Compute \(\hat q_t(\mathbf c)\) and \(S(t)\)
    using Eqs.~\ref{eq:timestep_sensitivity}
    and~\ref{eq:survival_ratio}
\EndFor

\State \Return sensitivity and survival-ratio curves
\end{algorithmic}
\end{algorithm}

\paragraph{Perturbation Survival Ratio.}
To see how the deviation between a baseline and its counterfactual changes over the rest of sampling, we compare its magnitude at the next model prediction with its magnitude in the final output. The replacement changes the sampled state \(\mathbf x_{t-1}\); processing this state at timestep \(t-1\) therefore provides the earliest comparison between the perturbed and baseline runs, before later denoising updates further modify their difference. For baseline run \(m\) and replacement \(r\), we measure this initial deviation as
\[
\Delta_{\mathrm{first}}^{(m,r)}(t)
=
\left\|
\mu_{\theta,m,r}^{\mathrm{pert}}(t-1)
-
\mu_{\theta,m}^{\mathrm{base}}(t-1)
\right\|_2,
\]
where \(\mu_{\theta,m}^{\mathrm{base}}(t-1)\) is the deterministic DDPM prediction from the baseline state, and  \(\mu_{\theta,m,r}^{\mathrm{pert}}(t-1)\) is the corresponding posterior mean predicted from the perturbed state, and their \(L_2\) distance measures the deviation in the first model prediction made after replacing \(\mathbf z_t\). Averaging over the \(M\) baseline runs and \(R\) replacements gives
\(\widehat{\Delta}_{\mathrm{first}}(t)\).

Let \(\widehat{\Delta}_{\mathrm{final}}(t)\) denote the corresponding mean deviation between the final outputs. We define
\begin{equation}
S(t)
=
\frac{\widehat{\Delta}_{\mathrm{final}}(t)}
     {\widehat{\Delta}_{\mathrm{first}}(t)+\epsilon},
\label{eq:survival_ratio}
\end{equation}
where \(\epsilon=10^{-12}\) prevents division by zero and is several orders of magnitude smaller than all observed first-response magnitudes. Values below \(1\) indicate that the baseline--counterfactual deviation decreases over denoising. Values near \(1\) indicate a similar final magnitude, while values above \(1\) indicate a larger final deviation.

\paragraph{Results: Counterfactual Diagnostic.}
\label{results: diagnostic}

\begin{table}[t]

\caption{Benchmarks evaluated in this study. Each column describes a fixed model--sampler--conditioning configuration.}

\label{tab:benchmarks}

\begin{center}

\setlength{\tabcolsep}{0.45mm}

\begin{tabular*}{\textwidth}{@{\extracolsep{\fill}}lccccc@{}}
\toprule
&
Maze2D
&
MPD-Prior
&
MPD-Guided
&
\begin{tabular}[c]{@{}c@{}}
PushT-\\
Transf.
\end{tabular}
&
PushT-CNN
\\
\midrule

Domain
& 2D Nav.
& Motion Plan
& Motion Plan
& visuomotor policy
& visuomotor policy
\\

Sampling object
& trajectory \(\boldsymbol\tau\)
& motion plan \(\boldsymbol\xi\)
& motion plan \(\boldsymbol\xi\)
& target-pos. \(\mathbf a_{1:H}\)
& target-pos. \(\mathbf a_{1:H}\)
\\

Diffused shape
& \(384 \times 6\)
& \(64 \times 14\)
& \(64 \times 14\)
& \(10 \times 2\)
& \(16 \times 2\)
\\

Architecture
& U-Net
& U-Net
& U-Net
& Transf.
& FiLM U-Net
\\

Training loss\textsuperscript{\(\ast\)}
& \(\mathcal L_{x_0}\)
& \(\mathcal L_{\epsilon,\mathcal H}\)
& \(\mathcal L_{\epsilon,\mathcal H}\)
& \(\mathcal L_{\epsilon}\)
& \(\mathcal L_{\epsilon}\)
\\

Prediction target
& \(\hat{\mathbf x}_{0,\theta}\)
& \(\hat{\boldsymbol\epsilon}_\theta\)
& \(\hat{\boldsymbol\epsilon}_\theta\)
& \(\hat{\boldsymbol\epsilon}_\theta\)
& \(\hat{\boldsymbol\epsilon}_\theta\)
\\

Modelled distribution
& \(p_\theta(\boldsymbol\tau)\)
& \(p_\theta(\boldsymbol\xi)\)
& \(p_\theta(\boldsymbol\xi)\)
& \(p_\theta(\mathbf a\mid\phi(\mathbf o_{1:2}))\)
& \(p_\theta(\mathbf a\mid\phi(\mathbf o_{1:2}))\)
\\

Sampling cond.\textsuperscript{\(\dagger\)}
& \(\mathcal H\)
& \(\mathcal H\)
& \(\mathcal H+G_{\rm cost}\)
& \(\phi(\mathbf o_{1:2})\)
& \(\phi(\mathbf o_{1:2})\)
\\

Noise schedule\textsuperscript{\(\ddagger\)}
& Cosine
& Exponential
& Exponential
& Cosine
& Cosine
\\

Denoising steps \(T\)
& 256
& 25
& 25
& 100
& 100
\\

\bottomrule
\end{tabular*}

\raggedright
\(\ast\)
\(\mathcal L_{x_0}
=
\mathbb E_i
\|\mathcal H\hat{\mathbf x}_{0,\theta}-\mathbf x_0\|_{1,w}\);
\(\mathcal L_{\epsilon,\mathcal H}
=
\mathbb E_{i,\boldsymbol\epsilon}
\|\mathcal H\hat{\boldsymbol\epsilon}_\theta-\boldsymbol\epsilon\|_2^2\);
\(\mathcal L_{\epsilon}
=
\mathbb E_{i,\boldsymbol\epsilon}
\|\hat{\boldsymbol\epsilon}_\theta-\boldsymbol\epsilon\|_2^2\).
\par

\(\dagger\) \(\mathcal H\): start--goal inpainting; \(G_{\rm cost}\): differentiable cost guidance; \(\phi(\mathbf o_{1:2})\): image/agent-position encoding.
\par

\(\ddagger\) Cosine: \(g_i=\cos^2\!\bigl(\frac{i/T+0.008}{1.008}\frac{\pi}{2}\bigr)\), \(\bar\alpha_i=g_i/g_0\), \(\beta_i=\min\{1-\bar\alpha_{i+1}/\bar\alpha_i,0.999\}\); exponential: \(\beta_i=10^{4i/(T-1)-4}\).
\par

\end{center}

\end{table}

We evaluate our counterfactual diagnostic on Maze2D~\citep{pmlr-v162-janner22a}, Motion Planning Diffusion (MPD)~\citep{carvalho2025motion}, and the PushT Diffusion policy~\citep{Chi2024Diffusion}. Maze2D uses start--goal trajectory inpainting. MPD is evaluated both with start--goal inpainting alone (MPD-P) and with additional differentiable cost guidance (MPD-G). Both PushT policies condition on encoded visual observations and agent position, using Transformer and FiLM U-Net architectures, respectively. Table~\ref{tab:benchmarks} describes these configurations in greater detail. Table~\ref{tab:perturbation_budget} reports our evaluation budgets used for stochastic averaging of each sensitivity profile. We probe all 25 MPD and 100 PushT denoising timesteps. For Maze2D, we probe a nonuniform subset of 114 timesteps from the 256-step chain and linearly interpolate the measured endpoint sensitivities onto the full integer timestep grid before constructing \(W_\alpha\). Here, a task is a distinct start--goal instance for Maze2D and MPD, or a closed-loop episode for PushT. Held-out intervention tasks are task instances not used to construct the aggregate diagnostic curve. For each configuration, the stochastic diversity windows are computed once from this curve using Equation~\ref{eq:capture-window} and applied unchanged, without per-task recalculation, to these held-out tasks.

Figure~\ref{fig:sensitivity-curves}(a) shows ten task-level endpoint-sensitivity curves per benchmark configuration: one for each distinct start--goal instance in Maze2D and MPD, and one for each closed-loop episode in PushT. Each curve is averaged over the baseline seeds and replacement draws in Equation~\ref{eq:timestep_sensitivity}.

The MPD-P and MPD-G profiles were nearly indistinguishable, indicating that late cost guidance did not materially alter the endpoint-sensitivity profile; therefore, only MPD-P is shown here. 
Sensitivity differs substantially across benchmarks in both range and location. Maze2D and MPD are most sensitive during early (noisy side) denoising, although Maze2D exhibits broader, more seed-variable peaks, MPD shows a consistent decay toward the end of denoising. In contrast, both PushT policies reach their highest sensitivity near the \textit{end} of denoising when the sample is  mostly clean, with a sharply late-dominated Transformer profile and greater episode-level heterogeneity for the CNN.

We additionally compared variation in capture-window length across stochastic baselines within a task with variation across tasks. For Maze2D and both PushT configurations, within-task variation was greater than between-task variation across all capture levels considered. MPD-P showed the opposite pattern, with substantially greater variation between tasks. Despite this task dependence, a single configuration-level mean profile was used unchanged to select intervention regions for held-out tasks in all benchmarks, and the resulting interventions remained effective. This suggests that the aggregate diagnostic can provide a useful configuration-level guide, while task-specific profiles may provide more precise intervention regions when additional diagnostic computation is justified. The full analysis is provided in Appendix~\ref{app:task_variation}.

These contrasting results highlight that the location and task dependence of stochastic sensitivity vary across model--sampler configurations. The stochastic diversity windows shown as dashed overlays in Figure~\ref{fig:sensitivity-curves}(a) summarize where the measured endpoint sensitivity is concentrated, whether that influence occurs early or late in denoising.

\begin{table}[t]

\caption{Evaluation budget used to construct the counterfactual diagnostic.
\(B\) denotes the number of samples generated in parallel for each timestep-noise perturbation repeat.}

\label{tab:perturbation_budget}

\begin{center}

\setlength{\tabcolsep}{3.5pt}

\begin{tabular}{@{}lccccc@{}}
\toprule
& Maze2D & \multicolumn{2}{c}{MPD} & \multicolumn{2}{c}{PushT} \\
\cmidrule(lr){3-4}
\cmidrule(lr){5-6}
& & Prior & Guided & Transf. & CNN \\
\midrule
Task conditions \(C\)        & 10  & 10  & 5  & 10  & 10 \\
Baseline samples/task \(M\)  & 10  & 10 & 10 & 10  & 10 \\
Timesteps probed \(T\)       & 114 & 25  & 25 & 100 & 100 \\
Repeats/timestep \(R\)       & 10  & 50  & 10 & 10  & 10 \\
Batch size \(B\)             & 50  & 50  & 10 & 1  & 1 \\
\bottomrule
\end{tabular}

\end{center}

\end{table}

\paragraph{Survival of Individual Perturbations.}
Figure~\ref{fig:sensitivity-curves}(b) reports the survival ratio \(S(t)\) from Equation~\ref{eq:survival_ratio} as \(\log_{10}S(t)\). The large immediate effect of stochastic perturbations at high-noise timesteps is expected from the diffusion schedule; more informative is how much of that perturbation remains consequential after subsequent denoising. PushT attenuates noisy-side perturbation responses by several orders of magnitude, leaving clean-side perturbations dominant where fewer denoising updates remain. MPD also strongly attenuates the initial response, but enough noisy-side influence survives to produce noisy-side endpoint dominance. Maze2D shows the strongest persistence, with some noisy-side perturbations approximately preserved or amplified.

Prior stability analyses show that reverse diffusion can progressively forget perturbations under suitable conditions, with earlier errors receiving increasing downstream attenuation~\citep{strasman2026forgetting}. Our results show that the extent of this attenuation differs substantially across model--sampler configurations. One possible explanation is that differences in the learned score field and conditioning constraints shape how rapidly perturbations lose influence, although SDW does not establish the underlying mechanism. Investigating this relationship across architectures, conditioning mechanisms, and sampler discretizations offers a direction for future work. These differences also demonstrate why influential timesteps must be identified empirically, motivating the selection of intervention intervals directly from the measured sensitivity profiles.

\section{Intervening on the Denoising Trajectory}
\label{sec:control}

The endpoint-sensitivity profiles in Figure~\ref{fig:sensitivity-curves} motivate two diagnostic-guided interventions. They reveal compact, model--sampler-specific regions of high sensitivity, defined by Equation~\ref{eq:capture-window}, alongside broad regions of comparatively low sensitivity. We therefore test whether targeting stochastic amplification to high-sensitivity regions increases feasible diversity and exposes further high-quality solutions, and whether the measured sensitivity profile can guide stochastic--deterministic timestep allocation to reduce network function evaluations (NFEs) while preserving performance.

\paragraph{Metrics.}
\label{para:metrics}
We report feasibility, feasible-sample cost, and feasible-sample diversity. Feasibility measures task completion under domain constraints; cost and diversity measure the quality and spread of the feasible outputs in each sampling batch. We denote the number of feasible outputs by \(F\).

For Maze2D and MPD, a trajectory is feasible if it reaches the goal without collision. For PushT, a closed-loop rollout is feasible if \(\max_s r_s \geq 0.999\), where \(r_s\) is the simulator reward at environment step \(s\).

For feasible trajectory \(i\), \(\mathbf{p}_{i,h}\) and \(\mathbf{v}_{i,h}\) denote position and velocity at waypoint \(h\), and \(a_i\) is the goal-proximity waypoint for Maze2D or the final waypoint for MPD. 
Its cost is:

\(J_i=\sum_{h=0}^{a_i-1}\|\mathbf{p}_{i,h+1}-\mathbf{p}_{i,h}\|_2
+\sum_{h=0}^{a_i-1}\|\mathbf{v}_{i,h+1}-\mathbf{v}_{i,h}\|_2\).
We calculate the mean feasible-sample cost as
\(
\label{eq:feasible-cost}
J=\frac{1}{F}\sum_{i=1}^{F}J_i .
\)
We additionally report the best feasible-sample cost \(J^*=\min_i J_i\). Both \(J\) and \(J^*\) are undefined when \(F=0\).

For PushT, \(J\) is the mean number of environment steps to first success, and \(J^*\) is the corresponding minimum over successful rollouts. Lower values indicate shorter, smoother trajectories or faster task completion.

Feasible-sample diversity is the mean pairwise distance between task-relevant output representations:
\begin{equation}
\label{eq:feasible-diversity}
D
=
\frac{2}{F(F-1)}
\sum_{i<j}
\left\|
\mathbf{y}_i-\mathbf{y}_j
\right\|_2 .
\end{equation}
For Maze2D and MPD, \(\mathbf{y}_i\) is the flattened trajectory. For PushT, it is the flattened block-pose trajectory, since success depends on the resulting block motion. Each block pose is represented as \((x,y,50\cos\theta,50\sin\theta)\) so that physically similar orientations remain close even when their angle values cross the \(-\pi/\pi\) boundary. Diversity \(D\) is undefined when \(F<2\).

For repeated evaluations, task-level metric means are computed over runs for which the corresponding metric is defined, and cross-task statistics are computed from these task-level means.

\subsection{Stochastic amplification}
\label{sec:stochastic_amplification}

\paragraph{Budgeted Target Interval \(I_{\alpha,K}\).}
The stochastic diversity window \(W_\alpha\) identifies the shortest contiguous interval containing an \(\alpha\) fraction of aggregate endpoint sensitivity where the sensitivity is a single-timestep measurement. Successively amplifying stochastic noise throughout \(W_\alpha\) could accumulate deviations across the window. Strength and temporal range of added sampling stochasticity is already known to affect sample quality~\citet{karras2022elucidating}. We therefore restrict amplification to a budget of \(K\) contiguous timesteps within the diagnosed high-sensitivity region. We treat \(K\) as an intervention-budget hyperparameter and evaluate a set of budgets for each sampler length. We denote the resulting intervention interval by \(I_K\), with \(|I_K|=K\). 

In essence, \(W_\alpha\) summarizes the diagnostic as a timestep window for capture level \(\alpha\), while \(K\) reflects a design choice for intervention within that window.

\paragraph{Amplification Mechanism.}
For a baseline sample \(m\), the stochastic component of the DDPM update is \(\boldsymbol{\eta}_t^{(m)}=\sigma_t\mathbf z_t^{(m)}\), where \(\mathbf z_t^{(m)}\sim\mathcal N(\mathbf 0,\mathbf I_d)\). Given intervention interval \(I_K\) and scale factor \(\lambda>1\), we modify sampler noise only in the selected budgeted interval:
\(
\label{eq:window-scaled-noise}
\tilde{\mathbf z}_t^{(\lambda,I_K,m)}
=
\begin{cases}
\lambda \mathbf z_t^{(m)}, & t\in I_K,\\
\mathbf z_t^{(m)}, & t\notin I_K.
\end{cases}
\)

We treat \(\lambda\) as an intervention-strength hyperparameter. To choose an initial range of values, we examined the ratio between the magnitude of the stochastic term and the model update across denoising. This gave a rough indication of when amplifying the noise would become substantial relative to the model update, rather than directly determining \(\lambda\). Example model/noise ratios are shown in Appendix~\ref{app:model_noise_ratio}. We then empirically tested a range of weaker to stronger amplification values.

For each intervention, the sampler is rerun with the task condition, initial latent, and all other noise draws fixed. For PushT, the same fixed intervention interval \(I_K\), identified from the configuration-level diagnostic, is applied independently within each diffusion policy call as the closed-loop episode replans.

\paragraph{Results.}
\label{sec:control_results}

\begin{table}[t]

\caption{Stochastic amplification results (abridged). $K$ is the number of denoising timesteps amplified and $\lambda$ the scalar multiplier applied to injected noise; $\lambda=1$ denotes the unmodified sampler. Values are mean $\pm$ SD across held-out tasks, showing task variability, except for Maze2D which shows one example task. The results show the feasibility--diversity trade-offs of various objectives: stronger amplification can increase diversity at some cost to feasibility, while noise attenuation can favour feasibility with lower diversity. Increased diversity can also improve best-of-batch sample quality whilst mean sample quality changes little or worsens.}

\label{tab:stochastic_scaling}

\begin{center}

\setlength{\tabcolsep}{7pt}

\begin{tabular}{@{}llrrrr@{}}
\toprule
Model & Amplification ($K$, $\lambda$) & Feasibility $F \uparrow$ & Diversity $D \uparrow$ & Mean cost $J \downarrow$ & Best cost $J^* \downarrow$ \\
\midrule

Maze2D
    & Baseline ($\lambda=1$)
    & $44$
    & $35.2$
    & $34.19$
    & $24.6$ \\
\cmidrule(lr){2-6}
    & $(5,0.75)$
    & $46$
    & $33.2$
    & $34.22$
    & $24.6$ \\
    & $(5,2.5)$
    & $37$
    & $44.3$
    & $33.45$
    & $18.1$ \\
    & $(5,3)$
    & $35$
    & $47.7$
    & $33.32$
    & $18.2$ \\

\midrule

MPD-P
    & Baseline ($\lambda=1$)
    & $19.40{\pm}21.16$
    & $4.75{\pm}1.50$
    & $12.76{\pm}4.38$
    & $11.22{\pm}4.49$ \\
\cmidrule(lr){2-6}
    & $(5,4)$
    & $20.20{\pm}15.77$
    & $9.89{\pm}0.82$
    & $14.12{\pm}3.03$
    & $10.91{\pm}3.58$ \\
    & $(10,4)$
    & $19.46{\pm}14.05$
    & $10.82{\pm}1.00$
    & $14.30{\pm}2.73$
    & $10.75{\pm}2.73$ \\

\addlinespace

MPD-G
    & Baseline ($\lambda=1$)
    & $47.77{\pm}5.12$
    & $6.62{\pm}3.06$
    & $13.98{\pm}3.78$
    & $9.80{\pm}2.03$ \\
\cmidrule(lr){2-6}
    & $(5,4)$
    & $46.72{\pm}5.71$
    & $10.92{\pm}2.11$
    & $14.70{\pm}2.15$
    & $9.41{\pm}1.31$ \\
    & $(10,4)$
    & $45.75{\pm}6.18$
    & $11.57{\pm}1.88$
    & $14.94{\pm}2.00$
    & $9.59{\pm}1.27$ \\

\midrule

PushT-T
    & Baseline ($\lambda=1$)
    & $17.2{\pm}13.6$
    & $352.0{\pm}211.4$
    & $204.13{\pm}39.44$
    & $144.33{\pm}44.75$ \\
\cmidrule(lr){2-6}
    & $(20,2)$
    & $15.3{\pm}11.9$
    & $528.7{\pm}442.6$
    & $196.65{\pm}49.86$
    & $148.44{\pm}61.82$ \\

\addlinespace

PushT-CNN
    & Baseline ($\lambda=1$)
    & $18.8{\pm}22.4$
    & $425.3{\pm}530.1$
    & $201.81{\pm}85.75$
    & $178.0{\pm}90.84$ \\
\cmidrule(lr){2-6}
    & $(20,2)$
    & $17.8{\pm}19.8$
    & $498.2{\pm}544.5$
    & $190.21{\pm}56.36$
    & $152.5{\pm}72.58$ \\

\bottomrule
\end{tabular}

\end{center}

\end{table}

Across the tested values for intervention length \(I_K\) and noise scale-factor \(\lambda\), stochastic amplification shows trade-offs across the performance metrics. Moderate amplification for example, can increase diversity within output batches with little loss of feasibility or sample quality, occasionally improving all three metrics. Stronger amplification however, often reduces the total number of collision-free samples in a batch and the batch-aggregated sample quality, while also resulting in a best-of-\(N\) single sample performance.

Table~\ref{tab:stochastic_scaling} reports selected \(K,\lambda\) combinations that allow comparison across configurations. The full set of tested combinations across all model configurations is provided in Appendix~\ref{app:full_stoch_amp_results}.

Figure~\ref{fig:mpd_qualitative_k5} provides qualitative examples of results for MPD-P and MPD-G. The paths shown are of the panda arm end-effector executing diffused motion samples within a 3D environment containing collision spheres, (spheres not shown for visual clarity). Within the sample batches where cost guidance is applied, we can see moderate stochastic amplification \(\lambda=3\) improves best-of-\(N\) sample quality, but also, most notably shows a distinct, previously unseen feasible route not observed with guidance alone nor by unguided amplification at any tested scale factor. This suggests that amplification opens access to new trajectory solutions, while cost guidance steers the resulting solutions toward feasibility. Explaining the mechanism remains future work. MPD-P (which is the same inpainting task, but without guidance), demonstrates a consistent improvement in diversity, and frequent improvement in the best sample of the batch, but often with feasibility and sample quality sacrifice at batch level aggregates. Further qualitative examples of all model configurations are provided in Appendix~\ref{app:stoch_amp_qualitative}.

\begin{figure}[h]
\centering
\includegraphics[width=0.49\textwidth]{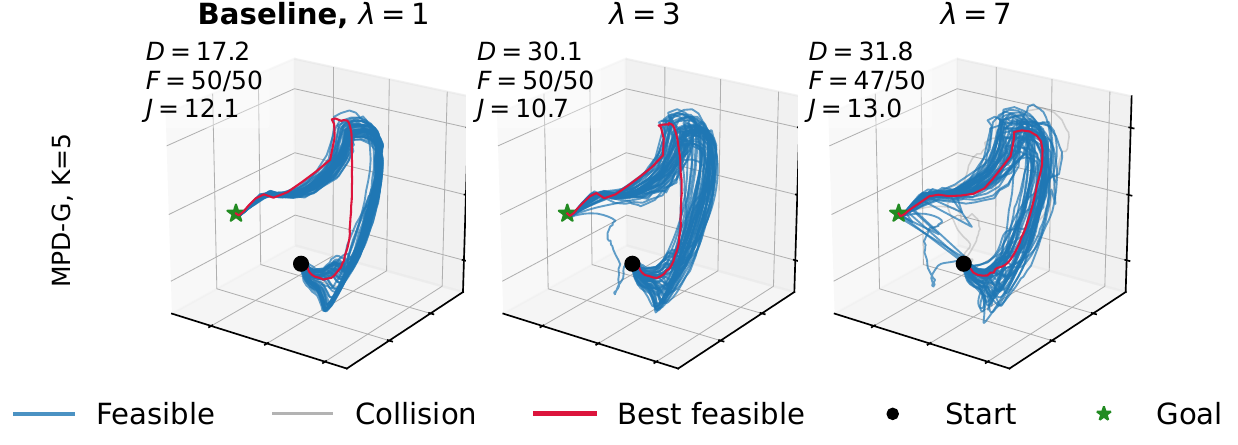}
\hfill
\includegraphics[width=0.49\textwidth]{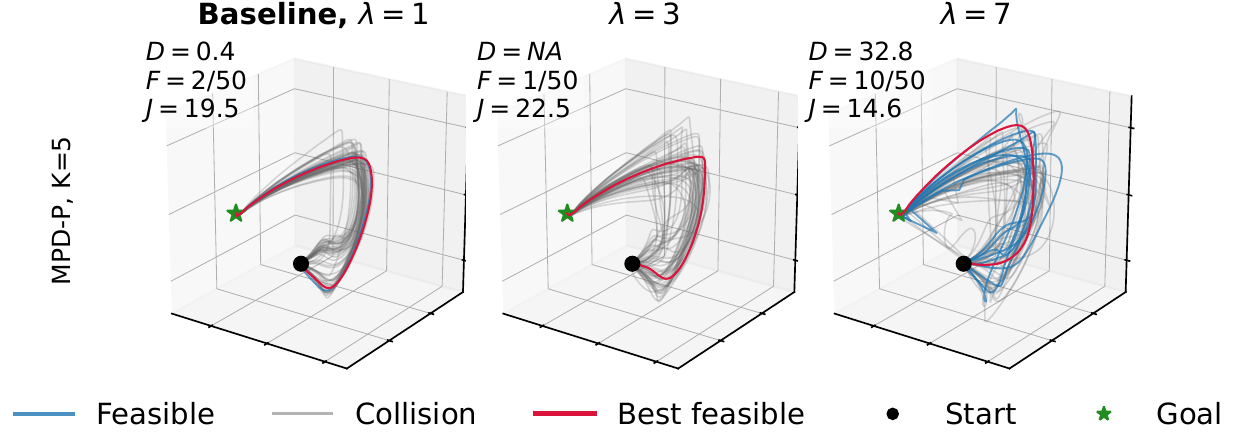}
\caption{Qualitative examples of stochastic amplification for MPD-G (left) and MPD-P (right). Plans show Panda arm end-effector trajectories; collision spheres are omitted for clarity. In MPD-G, combining amplification with guidance at \(\lambda=3\) reveals an additional route, while stronger amplification at \(\lambda=7\) shows diminishing returns. In MPD-P, amplification increases diversity and the number of collision-free trajectories.}
\label{fig:mpd_qualitative_k5}
\end{figure}

\paragraph{Stochastic Amplification Ablation Study.}
Because the diagnostic and intervention are distinct mechanisms, a crucial question is whether the diversity gains are a direct result from targeting the diagnosed location or simply from the act of injecting higher-magnitude noise into the denoising chain. This is particularly important when the diagnostic reveals opposite sensitivity regimes across benchmarks as we have seen: MPD concentrates influence near the start of denoising, while PushT concentrates it near the end. To test this, we apply the same \((K,\lambda)\) stochastic amplification budget to the diagnosed window, and also to a control region in the middle of denoising, and a control region at the opposite end of the targeted \(I_K\). Table~\ref{tab:amplification-location-ablation} shows that the diagnosed window produces the largest mean gains in feasible-sample diversity \(\Delta D\) across both MPD and PushT. Although the gain magnitude is task-dependent, only the diagnosed interval produces a substantial mean diversity increase; the control intervals yield small or inconsistent changes.

\begin{table}[t]

\caption{Intervention-target ablation for stochastic amplification. We report feasible-sample diversity change \(\Delta D\) (mean \(\pm\) SD) for the diagnosed interval \(I_{\alpha,K}\), a middle interval, and an opposite-side interval.}

\label{tab:amplification-location-ablation}

\begin{center}

\setlength{\tabcolsep}{1.5pt}

\begin{tabular}{@{}lcccc@{}}
\toprule
& & \multicolumn{3}{c}{Feasible-sample diversity change \(\Delta D \uparrow\)} \\
\cmidrule(lr){3-5}
Model & \((K,\lambda)\) & Diagnosed (Ours) & Middle & Opposite \\
\midrule

MPD-G
& \((5,4)\)
& \(\mathbf{+4.30 \pm 1.48}\)
& \(+0.39 \pm 0.20\)
& \(+0.03 \pm 0.01\)
\\

MPD-P
& \((5,4)\)
& \(\mathbf{+5.05 \pm 1.59}\)
& \(+0.29 \pm 0.55\)
& \(-0.24 \pm 0.81\)
\\

PushT-T
& \((10,2)\)
& \(\mathbf{+194.5 \pm 220.9}\)
& \(+1.3 \pm 70.6\)
& \(+2.1 \pm 87.9\)
\\

\bottomrule
\end{tabular}

\end{center}

\end{table}

\subsection{DDPM-DDIM Hybrid Sampling}
\label{sec: SDW-Hybrid}

The same diagnosed intervals given earlier can also be used to reduce sampling cost. We construct a DDPM--DDIM hybrid, which we call \textit{SDW-hybrid}, which retains DDPM posterior sampling on diagnosed high-sensitivity timesteps \(I_K\), while traversing lower-sensitivity regions with deterministic DDIM skips. Related hybrid DDPM--DDIM sampling has been explored previously in HybridDPS, where the transition between deterministic and stochastic sampling is selected empirically as a fixed global hyperparameter~\citep{liu2026hybriddps}. In contrast, \textit{SDW-hybrid} uses the measured endpoint-sensitivity profile to determine where stochastic DDPM transitions are retained. In MPD-G, we additionally preserve DDPM transitions in the late guidance region to match the full-DDPM guided baseline.

For \(t\in I_K\), we use the original DDPM posterior-sampling transition~\citep{ho2020denoising},
\[
\mathbf x_{t-1}
=
\boldsymbol{\mu}_\theta(\mathbf x_t,t,\mathbf c)
+
\sigma_t\mathbf z_t,
\qquad
\mathbf z_t\sim\mathcal N(\mathbf 0,I).
\]
In lower-sensitivity regions, let \(u<t\) denote the next retained timestep on the clean side of the reverse process. We skip directly from \(t\) to \(u\) using the deterministic DDIM update~\citep{song2020denoising},
\[
\mathbf x_u
=
\sqrt{\bar\alpha_u}\,
\hat{\mathbf x}_{0,\theta}(\mathbf x_t,t,\mathbf c)
+
\sqrt{1-\bar\alpha_u}\,
\boldsymbol{\epsilon}_\theta(\mathbf x_t,t,\mathbf c).
\]

For MPD-P, the hybrid sampler retains \(K\) consecutive DDPM transitions in the diagnosed high-sensitivity region and uses stride-5 DDIM updates elsewhere. For MPD-G, the late guided DDPM region is also retained. DDIM shortcuts are shortened where necessary so that they do not skip a retained DDPM transition. The specific timestep schedules used are provided in Appendix~\ref{app:SDWHybrid_steps}.

\subsection{Results: SDW-Hybrid Sampler}

\begin{figure}[ht]
\centering
\includegraphics[width=0.49\textwidth]{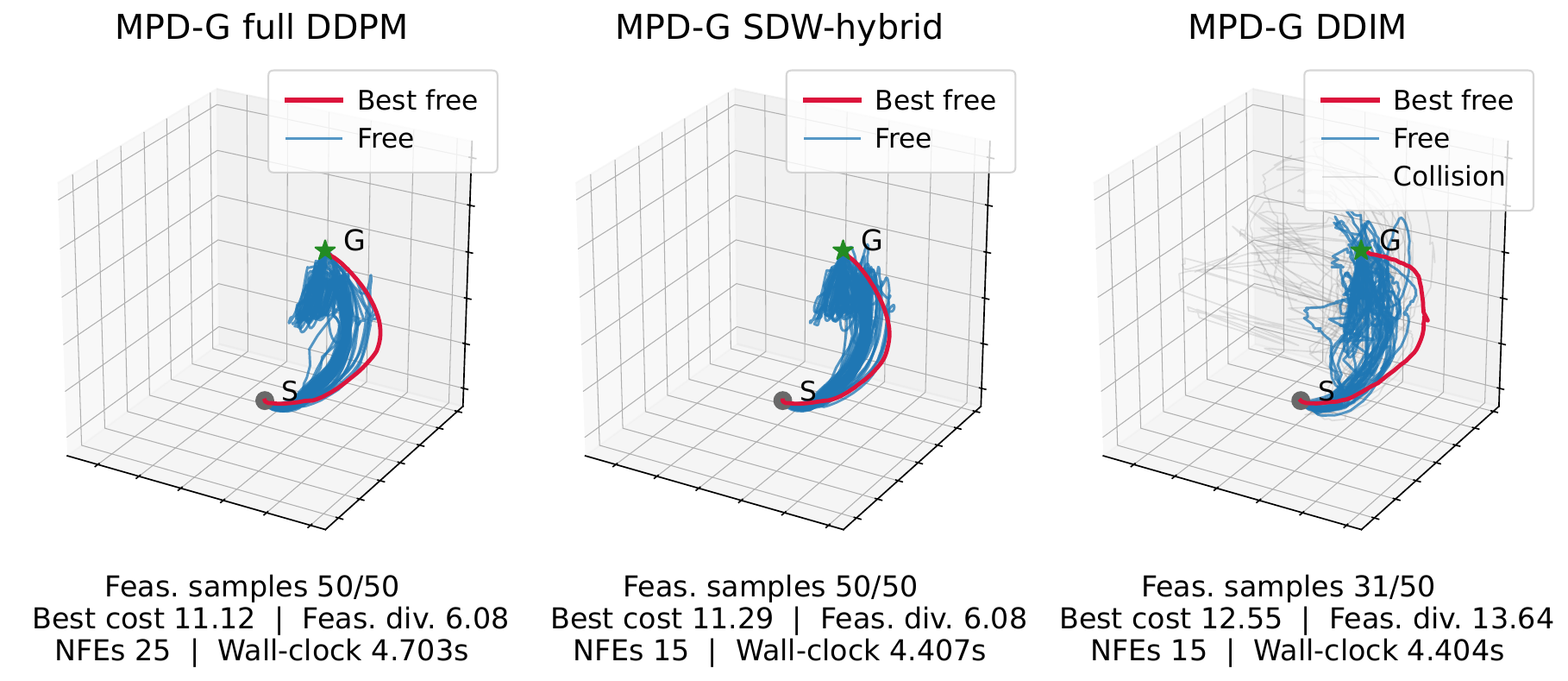}%
\hfill
\includegraphics[width=0.49\textwidth]{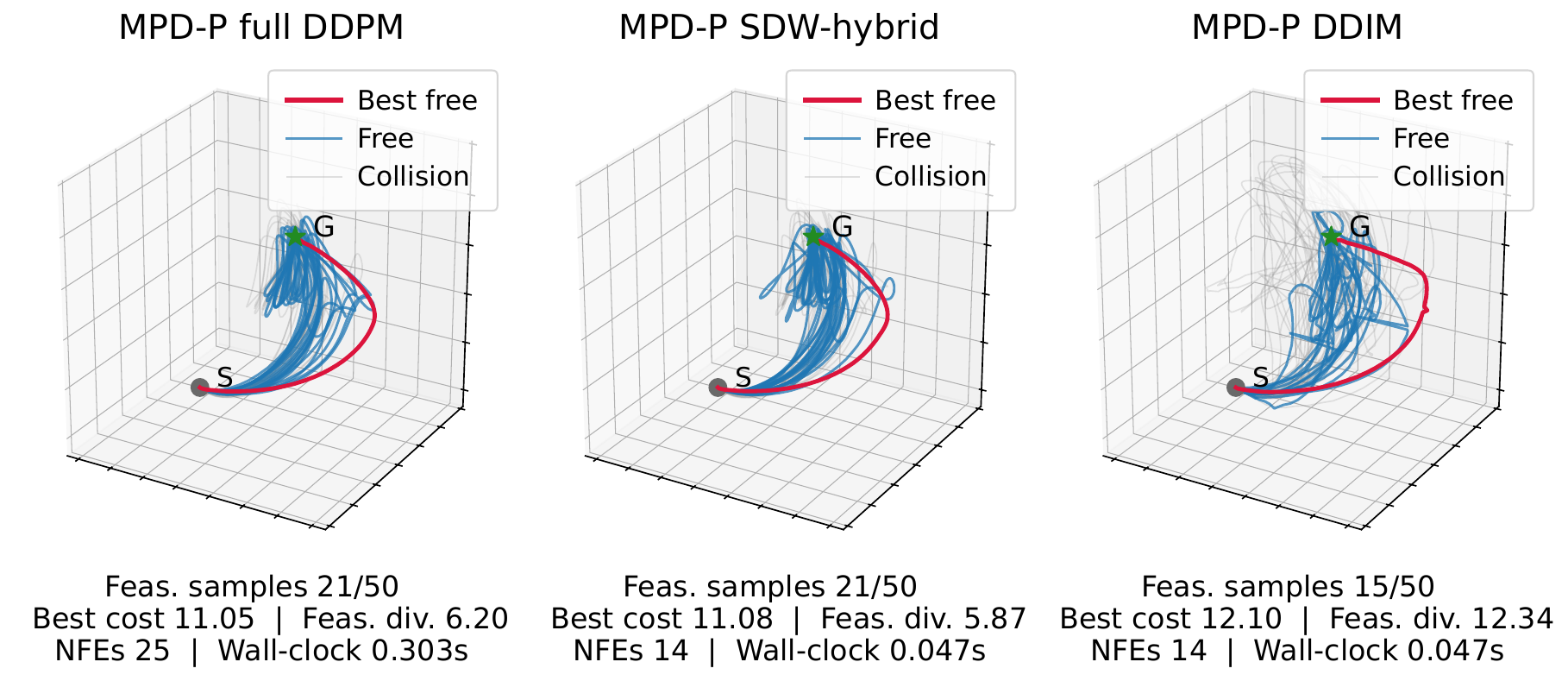}
\caption{Qualitative MPD paths for SDW-hybrid: full DDPM, SDW-hybrid, and DDIM for the same task and seed. Left shows inpainting + guided-sampling, right shows inpainting only tasks. SDW-hybrid produces comparable feasible trajectory quality with fewer denoiser evaluations and lower wall-clock time.}

\label{fig:sdw_hybrid_qual}
\end{figure}

Table~\ref{tab:sdw_hybrid} compares SDW-hybrid with the original full-DDPM sampler over six held-out tasks, with 50 run seeds and 50 samples per seed. Replacing low-sensitivity regions with DDIM skips substantially reduces denoiser evaluations while leaving the main task metrics close to the full sampler. For MPD-P, \(K=5\) and \(K=10\) reduce NFEs by \(60\%\) and \(44\%\), with wall-clock reductions of approximately \(85\%\) and \(79\%\). For MPD-G, NFEs fall by \(40\%\), but wall-clock time decreases by only \(5\%\) because the unchanged guidance computation dominates runtime.
Figure~\ref{fig:sdw_hybrid_qual} shows representative trajectory batches for MPD-G and MPD-P. In both cases, SDW-hybrid produces a similar set of feasible trajectories to full DDPM, with comparable best cost and feasible-sample diversity. The deterministic DDIM samples are substantially more dispersed, but this wider spread is accompanied by more collisions and fewer feasible trajectories.

To test whether these differences arise specifically from the placement of stochastic transitions, Table~\ref{tab:hybrid_timestep_ablation} compares SDW-hybrid with a matched middle-window hybrid. The control retains exactly the same NFE budget and DDPM/DDIM composition as SDW-hybrid, but moves the retained stochastic DDPM transitions from the diagnosed high-sensitivity region to an intermediate region of the reverse process. This is intended as a targeted location control, rather than an exhaustive sweep over alternative placements: the equal-NFE DDIM comparison in Table~\ref{tab:sdw_hybrid} separately controls for network-evaluation budget, while this ablation isolates whether retaining stochastic transitions in the diagnosed region is consequential.

Moving the retained DDPM transitions away from the diagnosed high-sensitivity region reduces performance. For MPD-P, feasibility falls from \(21/50\) to \(12/50\) for \(K=5\) and to \(14/50\) for \(K=10\), with higher feasible-sample cost in both cases. For MPD-G, feasibility falls from \(45/50\) to \(34/50\), while cost increases from \(9.97\) to \(11.09\). The middle-window variants produce more diverse feasible trajectories, but with lower feasibility and higher cost. This shows that where stochastic transitions are retained matters, and supports using the measured sensitivity profile to choose their location.

For MPD-P, the full-DDPM runtime shown in Table~\ref{tab:sdw_hybrid} summarizes both \(K\) experiments; statistical comparisons use the corresponding full-DDPM baseline from the same experiment.
Runtime reductions were consistent across tasks. For the bootstrap analysis, each SDW-hybrid result is compared with the corresponding full-DDPM result for the same task and run seed. Task-cluster bootstrap 95\% intervals excluded zero for MPD-G (\(\Delta=-0.210\)s, CI \([-0.215,-0.204]\)), MPD-P with \(K=5\) (\(\Delta=-0.195\)s, CI \([-0.202,-0.187]\)), and MPD-P with \(K=10\) (\(\Delta=-0.170\)s, CI \([-0.180,-0.160]\)). In contrast, the corresponding intervals for feasibility, cost, and diversity spanned zero, indicating no consistent directional change in these metrics across tasks.

\begin{table}[t]

\caption{SDW-hybrid sampling on MPD compared with full DDPM and deterministic DDIM. Values are mean \(\pm\) SD across six held-out tasks with 50 run seeds per task; parentheses report relative change from the corresponding full-DDPM baseline. Best-cost and diversity values are calculated only for runs where those metrics are defined.}

\label{tab:sdw_hybrid}

\begin{center}

\setlength{\tabcolsep}{1.3pt}

\begin{tabular}{@{}lccccc@{}}
\toprule
Sampler
& NFEs \(\downarrow\)
& Wall-clock (s) \(\downarrow\)
& Feasibility/50 \(\uparrow\)
& Best Cost \(\downarrow\)
& Diversity \(\uparrow\)
\\
\midrule

\multicolumn{6}{@{}l}{\textit{MPD-P (inpainting only)}} \\

Full DDPM
& 25
& \(0.222{\pm}0.075\)
& \(20{\pm}19\)
& \(9.88{\pm}2.72\)
& \(4.64{\pm}2.01\)
\\

SDW-hybrid, \(K=5\)
& 10
& \(0.033{\pm}0.000\) (-85\%)
& \(21{\pm}18\) (+5\%)
& \(9.86{\pm}2.65\) (0\%)
& \(5.11{\pm}1.94\) (+10\%)
\\

DDIM, \(K=5\)
& 10
& \(0.033{\pm}0.001\) (-85\%)
& \(12{\pm}10\) (-38\%)
& \(13.09{\pm}3.54\) (+32\%)
& \(13.06{\pm}1.96\) (+182\%)
\\

SDW-hybrid, \(K=10\)
& 14
& \(0.046{\pm}0.001\) (-79\%)
& \(21{\pm}19\) (+4\%)
& \(10.08{\pm}3.13\) (+2\%)
& \(4.74{\pm}1.60\) (+2\%)
\\

DDIM, \(K=10\)
& 14
& \(0.046{\pm}0.001\) (-79\%)
& \(16{\pm}11\) (-21\%)
& \(11.46{\pm}3.36\) (+16\%)
& \(11.60{\pm}1.76\) (+150\%)
\\

\midrule

\multicolumn{6}{@{}l}{\textit{MPD-G (inpainting + cost guidance)}} \\

Full DDPM
& 25
& \(4.614{\pm}0.080\)
& \(46{\pm}7\)
& \(9.86{\pm}2.00\)
& \(6.52{\pm}2.65\)
\\

SDW-hybrid, \(K=5\)
& 15
& \(4.405{\pm}0.009\) (-5\%)
& \(45{\pm}7\) (-2\%)
& \(9.97{\pm}2.11\) (+1\%)
& \(6.53{\pm}2.46\) (0\%)
\\

DDIM, \(K=5\)
& 15
& \(4.406{\pm}0.007\) (-5\%)
& \(27{\pm}9\) (-40\%)
& \(11.68{\pm}1.98\) (+18\%)
& \(13.65{\pm}1.57\) (+109\%)
\\

\bottomrule
\end{tabular}

\end{center}

\end{table}

\begin{table}[ht]

\caption{Timestep-selection ablation for SDW-hybrid. The middle-window control uses the same number of network evaluations and the same DDPM/DDIM composition as SDW-hybrid, but moves the retained stochastic DDPM steps away from the diagnosed high-sensitivity region. Values are mean \(\pm\) SD over 50 run seeds for each of six held-out tasks.}

\label{tab:hybrid_timestep_ablation}

\begin{center}

\setlength{\tabcolsep}{1.3pt}

\begin{tabular}{@{}llcccc@{}}
\toprule
Setting
& Placement
& NFEs (DDPM/DDIM)
& Feasibility/50 \(\uparrow\)
& Cost \(\downarrow\)
& Diversity \(\uparrow\)
\\
\midrule

MPD-P, \(K=5\)
& SDW
& 10 (5/5)
& \(21{\pm}18\)
& \(9.86{\pm}2.65\)
& \(5.11{\pm}1.94\)
\\

MPD-P, \(K=5\)
& Middle
& 10 (5/5)
& \(12{\pm}10\)
& \(13.08{\pm}3.11\)
& \(13.37{\pm}1.81\)
\\

\midrule

MPD-P, \(K=10\)
& SDW
& 14 (10/4)
& \(21{\pm}19\)
& \(10.08{\pm}3.13\)
& \(4.74{\pm}1.60\)
\\

MPD-P, \(K=10\)
& Middle
& 14 (10/4)
& \(14{\pm}11\)
& \(12.37{\pm}3.02\)
& \(12.55{\pm}1.82\)
\\

\midrule

MPD-G, \(K=5\)
& SDW
& 15 (12/3)
& \(45{\pm}7\)
& \(9.97{\pm}2.11\)
& \(6.53{\pm}2.46\)
\\

MPD-G, \(K=5\)
& Middle
& 15 (12/3)
& \(34{\pm}10\)
& \(11.09{\pm}1.74\)
& \(13.45{\pm}1.41\)
\\

\bottomrule
\end{tabular}

\end{center}

\end{table}

\section{Conclusion}
\label{sec:conclusion}
We introduced a counterfactual diagnostic pipeline that measures where injected stochasticity influences the final output and identifies compact \textit{stochastic diversity windows} from the resulting sensitivity profiles. Across robotic applications, sensitivity profiles differed sharply across model--sampler--conditioning configurations, highlighting the value of empirical diagnosis for locating influential denoising regions. The resulting windows transferred to held-out tasks and supported two complementary interventions. Targeted stochastic amplification increased feasible-sample diversity and exposed new high-quality solution families; in MPD-G, combining it with cost guidance revealed a distinct feasible route not observed with either mechanism alone. \textit{SDW-hybrid} reduced denoiser evaluations and wall-clock time while feasibility, cost, and diversity remained broadly similar across held-out tasks. More broadly, we believe SDW provides a practical toolkit for probing denoising dynamics.

\section{Limitations and Future Work}
The diagnostic requires source-level sampler access to store, replace, and replay per-timestep noise draws, and incurs significant upfront computational cost for accuracy from evaluating each timestep across multiple baseline seeds and replacement draws. This cost is paid once for a given frozen pretrained model and sampler, after which the sensitivity curves and intervention windows can be reused across tasks as we have demonstrated. Our study focuses on a range of domain applications for robotics, extending the approach to other generative diffusion model domains is an area for future work.
The diagnostic shows where stochastic perturbations contract, persist, or amplify, but not why nor how these behaviors occur, we feel the diagnostic can serve as a toolkit for future research to support mechanistic study of denoising dynamics.

\paragraph{Use of Generative AI.}
ClaudeAI was used for language polishing and LaTeX formatting. The authors verified all output and have full responsibility for the manuscript.

\bibliography{Diversity_windows_paper}
\bibliographystyle{tmlr}

\appendix

\section{Unbiased Monte Carlo Estimator}
\label{app:mc_unbiased}

We formalize the population quantity estimated by
Equation~\ref{eq:timestep_sensitivity} in the main paper. Fix a task condition
\(\mathbf c\) and denoising timestep \(t\). Let
\((\mathbf x_T,\mathbf z_{1:T})\) denote the baseline sampler randomness,
let \(\mathbf z'_t\) be an independent replacement draw from the same
distribution as \(\mathbf z_t\), and write
\(\mathbf z_{1:T}^{(t\leftarrow\mathbf z'_t)}\) for the noise sequence
obtained by replacing only \(\mathbf z_t\). The population timestep-level
endpoint sensitivity is
\begin{equation}
q_t(\mathbf c)
=
\mathbb{E}
\left[
\Delta\!\left(
S_\theta(\mathbf c,\mathbf x_T,\mathbf z_{1:T}),
S_\theta\!\left(
\mathbf c,
\mathbf x_T,
\mathbf z_{1:T}^{(t\leftarrow\mathbf z'_t)}
\right)
\right)
\right],
\label{eq:population_timestep_sensitivity}
\end{equation}
where the expectation is over the baseline sampler randomness and the
independent replacement draw.

\begin{proposition}
\label{prop:mc_estimator}
Assume that the baseline draws
\((\mathbf x_T^{(m)},\mathbf z_{1:T}^{(m)})\) are independent and identically
distributed across \(m=1,\ldots,M\). For each baseline \(m\), let
\(\mathbf z_t'^{(m,r)}\), \(r=1,\ldots,R\), be independent replacement draws
from the same distribution as \(\mathbf z_t\), independent of the
corresponding baseline draw. Then
\begin{equation}
\hat q_t(\mathbf c)
=
\frac{1}{MR}
\sum_{m=1}^{M}
\sum_{r=1}^{R}
\Delta\!\left(
\mathbf x_0^{(\mathbf c,m)},
\tilde{\mathbf x}_0^{(\mathbf c,m,t,r)}
\right)
\label{eq:appendix_mc_estimator}
\end{equation}
is an unbiased Monte Carlo estimator of \(q_t(\mathbf c)\). If the deviation
has finite variance, then \(\hat q_t(\mathbf c)\) is consistent as
\(M\to\infty\) for any fixed \(R\geq1\).
\end{proposition}

\begin{proof}
Let
\[
Y_{m,r}
=
\Delta\!\left(
\mathbf x_0^{(\mathbf c,m)},
\tilde{\mathbf x}_0^{(\mathbf c,m,t,r)}
\right).
\]
By construction, each \(Y_{m,r}\) is obtained from a baseline draw by
replacing only its timestep-\(t\) noise with an independent draw. Therefore,
\[
\mathbb{E}[Y_{m,r}]
=
q_t(\mathbf c).
\]
By linearity of expectation,
\[
\mathbb{E}[\hat q_t(\mathbf c)]
=
\frac{1}{MR}
\sum_{m=1}^{M}
\sum_{r=1}^{R}
\mathbb{E}[Y_{m,r}]
=
q_t(\mathbf c),
\]
which proves unbiasedness.

Repeats sharing a baseline need not be independent because they reuse
\(\mathbf x_T^{(m)}\) and the unchanged components of
\(\mathbf z_{1:T}^{(m)}\). Define the baseline-level average
\[
A_m
=
\frac{1}{R}
\sum_{r=1}^{R}
Y_{m,r}.
\]
The \(A_m\) are independent and identically distributed across \(m\), with
\(\mathbb{E}[A_m]=q_t(\mathbf c)\) and finite variance. Hence, by the law of
large numbers,
\[
\hat q_t(\mathbf c)
=
\frac{1}{M}
\sum_{m=1}^{M}
A_m
\xrightarrow[M\to\infty]{p}
q_t(\mathbf c),
\]
which proves consistency.
\end{proof}

\paragraph{PushT.}
For PushT, the selected replans within an episode have different fixed
observation-conditioning contexts
\(\mathbf c_1,\ldots,\mathbf c_M\). Conditional on these contexts, define the
episode-level population sensitivity as
\[
q_t(\mathbf c)
=
\frac{1}{M}
\sum_{m=1}^{M}
q_t(\mathbf c_m),
\]
where \(q_t(\mathbf c_m)\) is the expected endpoint sensitivity for the
diffusion policy call at replan \(m\). Equation~\ref{eq:timestep_sensitivity}
averages the corresponding Monte Carlo estimates across the selected
replans. Since each local estimate is unbiased for \(q_t(\mathbf c_m)\),
linearity of expectation gives
\[
\mathbb{E}\!\left[
\hat q_t(\mathbf c)
\mid
\mathbf c_1,\ldots,\mathbf c_M
\right]
=
\frac{1}{M}
\sum_{m=1}^{M}
q_t(\mathbf c_m)
=
q_t(\mathbf c).
\]
Thus, for PushT, Equation~\ref{eq:timestep_sensitivity} is unbiased for the
mean endpoint sensitivity across the selected replanning contexts in that
episode.

\section{Task-Level Variation in Capture Windows}
\label{app:task_variation}

\begin{figure}[p]
    \centering

    \includegraphics[
        width=0.86\textwidth
    ]{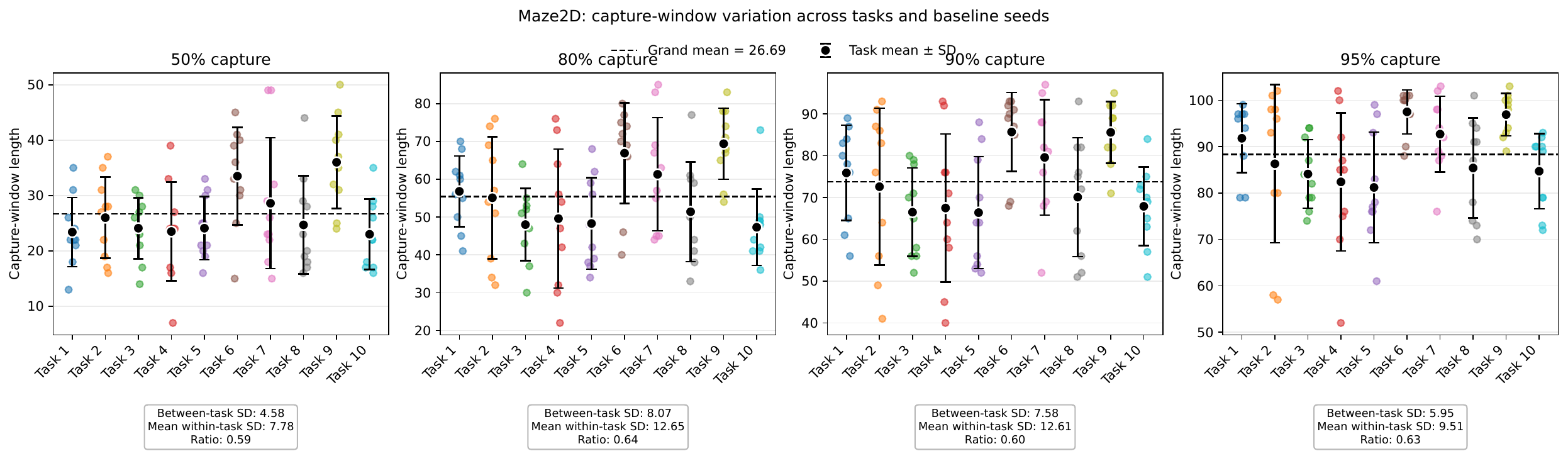}
    \caption{Maze2D capture-window variation across tasks and baseline seeds.}
    \label{fig:task_variation_maze2d}

    \vspace{0.4em}

    \includegraphics[
        width=0.86\textwidth
    ]{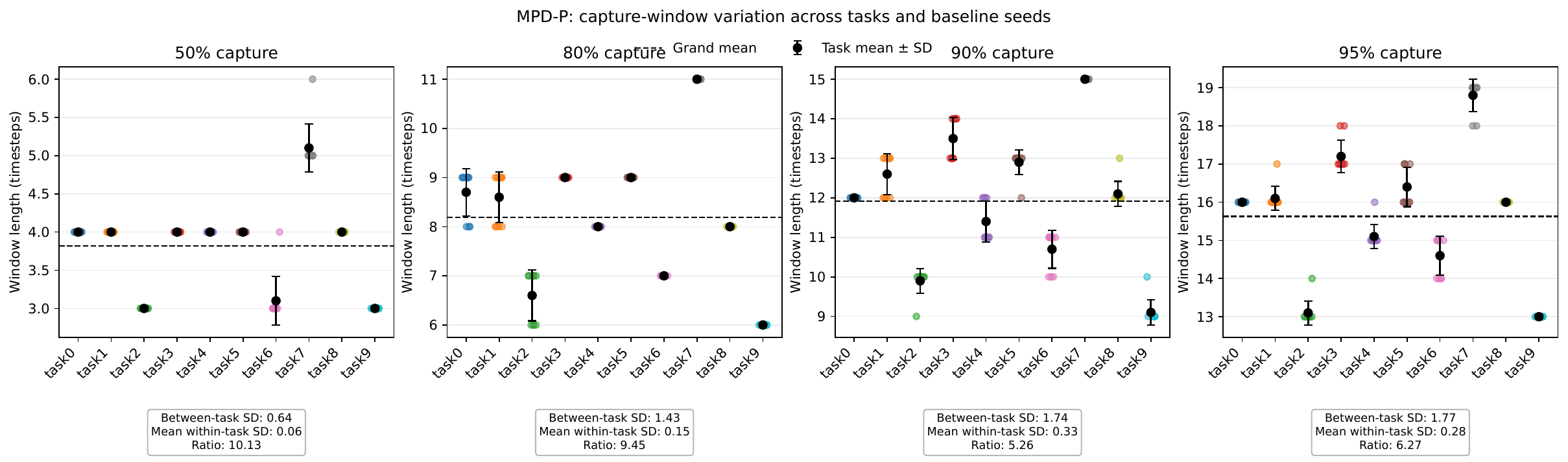}
    \caption{MPD-P capture-window variation across tasks and baseline seeds.}
    \label{fig:task_variation_mpd}
\end{figure}

\begin{figure}[p]
    \centering

    \includegraphics[
        width=0.86\textwidth
    ]{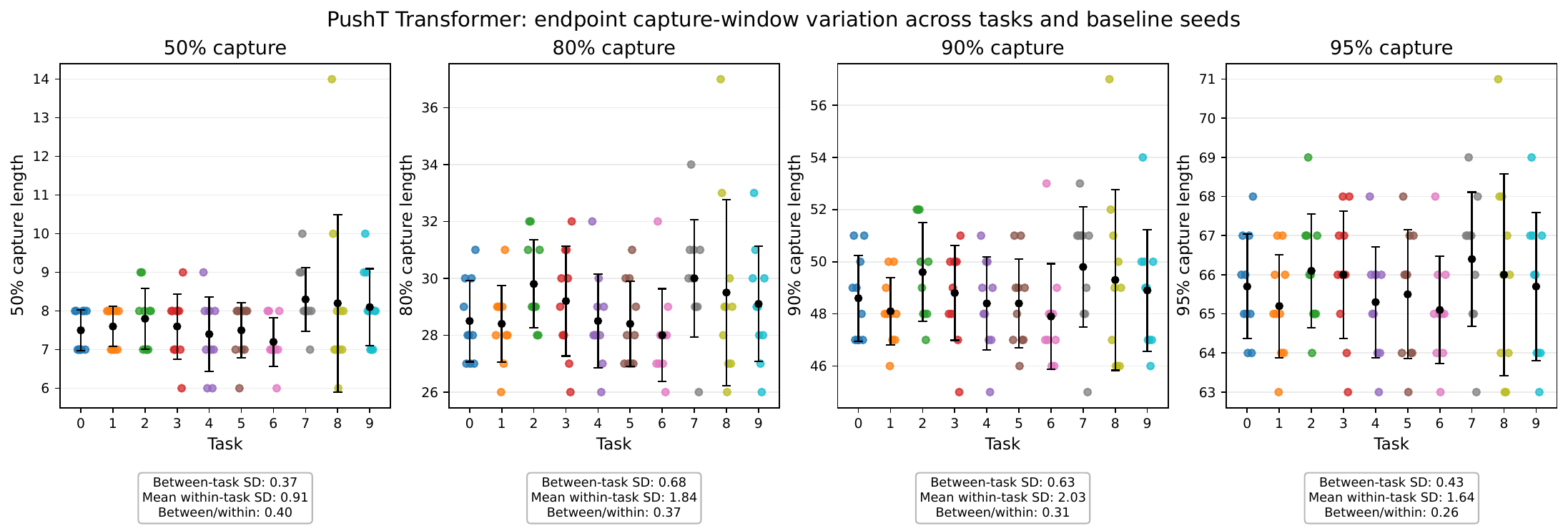}
    \caption{PushT Transformer capture-window variation across tasks and baseline seeds.}
    \label{fig:task_variation_pusht_transformer}

    \vspace{0.4em}

    \includegraphics[
        width=0.86\textwidth
    ]{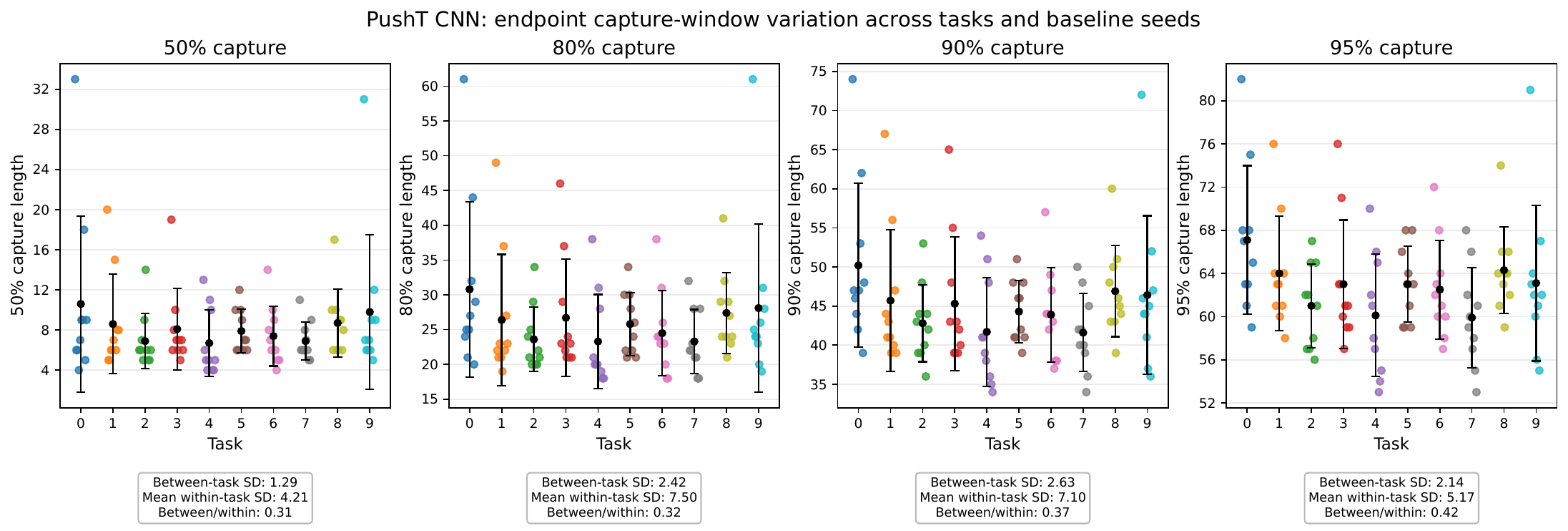}
    \caption{PushT CNN capture-window variation across tasks and baseline seeds.}
    \label{fig:task_variation_pusht_cnn}
\end{figure}

As discussed within the results of section~\ref{results: diagnostic}, for each model--sampler configuration, we examined how the capture-window length varies across tasks relative to stochastic variation within a task. For each task, the coloured points in Figures~\ref{fig:task_variation_maze2d}--\ref{fig:task_variation_pusht_cnn} show capture-window lengths obtained from individual baseline stochastic realisations, while the black markers show the task mean and standard deviation.

We summarize the relative sources of variation using
\[
R_\alpha
=
\frac{
\operatorname{SD}_{\mathrm{tasks}}
\left(
\overline{|W_\alpha|}
\right)
}{
\operatorname{mean}_{\mathrm{tasks}}
\left[
\operatorname{SD}_{\mathrm{baseline}}
\left(
|W_\alpha|
\right)
\right]
},
\]
where \(|W_\alpha|\) is the length of the shortest contiguous window capturing fraction \(\alpha\) of endpoint sensitivity. Values \(R_\alpha<1\) indicate that variation across stochastic baselines within a task exceeds variation between task means, whereas \(R_\alpha>1\) indicates greater task-to-task variation.

Maze2D and both PushT configurations have \(R_\alpha<1\) at every capture level considered, indicating that their task-to-task differences are smaller than the stochastic variation observed within individual tasks. MPD-P exhibits the opposite pattern, with \(R_\alpha\) ranging from approximately \(5.3\) to \(10.1\), showing substantially greater variation between tasks than within them.

Despite these differences, the interventions in the main paper use a single configuration-level sensitivity profile, obtained by averaging across diagnostic tasks, and apply the resulting intervention regions unchanged to held-out tasks. The intervention results therefore show that a configuration-level diagnostic can provide a useful common guide even when task-level window locations vary. At the same time, the stronger task dependence observed for MPD-P suggests that task-specific diagnostics may provide more precise intervention regions when additional diagnostic computation is justified.

\FloatBarrier
\section{Model Update to Noise Magnitude Ratios}
\label{app:model_noise_ratio}

To inform on starting points for appropriate scale-factors for the stochastic amplification experiments of section~\ref{sec:stochastic_amplification}, we examined the magnitude ratios of the deterministic model update to the per-step noise injections (of the unmodified null baseline).

We find a wide variation of ratios across the different model configurations, and that the ratios are stable across different tasks within each model configuration. But we also found that although useful as a starting point, a set of useful trade-off scale factors are found empirically.

\begin{figure*}[h]
    \centering
    \begin{minipage}[h]{0.48\textwidth}
        \centering
        \includegraphics[
            width=\linewidth
        ]{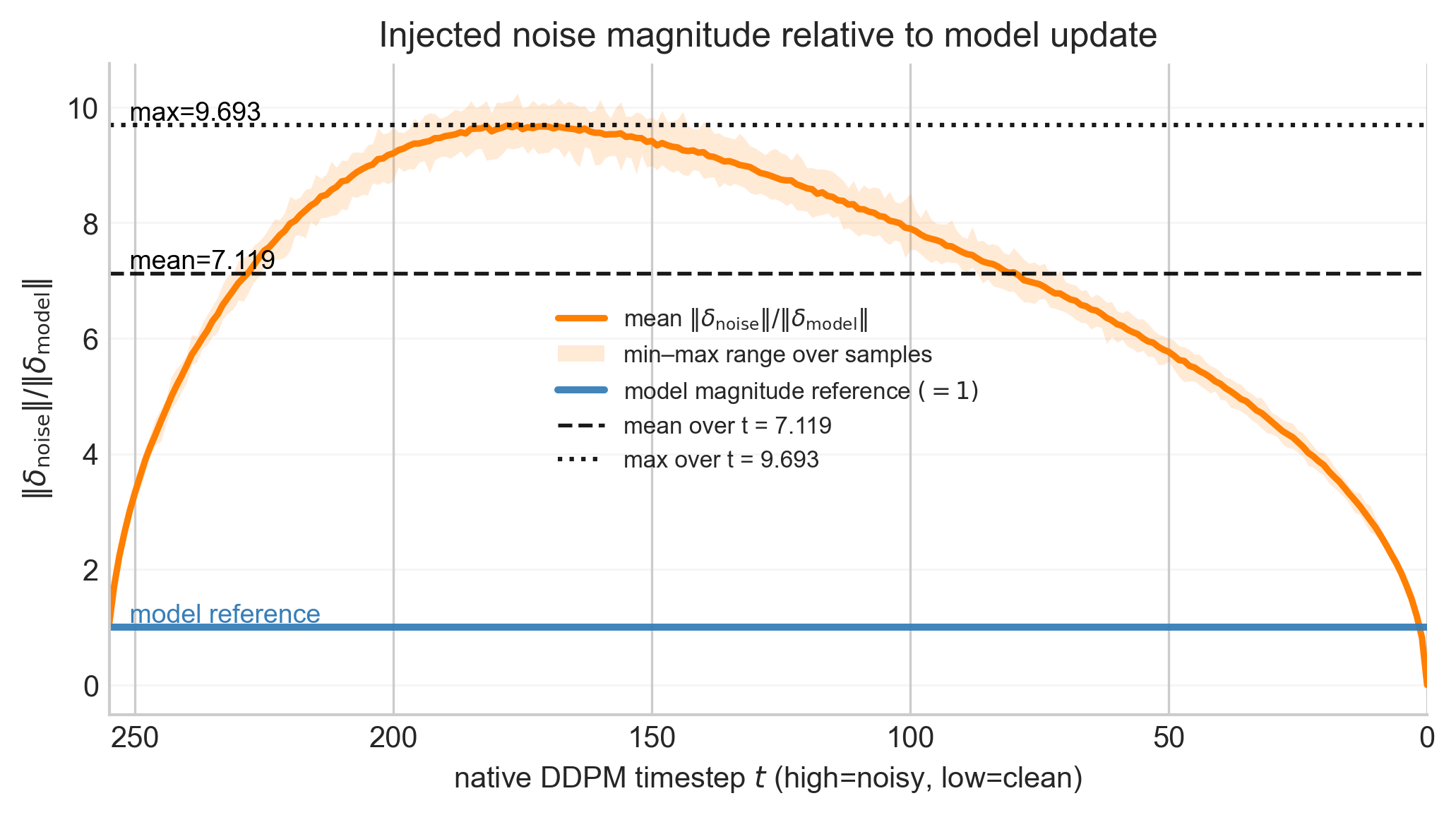
        }
        \vspace{0.3em}
        \textbf{(a) Maze2D}
    \end{minipage}
    \hfill
    \begin{minipage}[h]{0.48\textwidth}
        \centering
        \includegraphics[width=\linewidth]{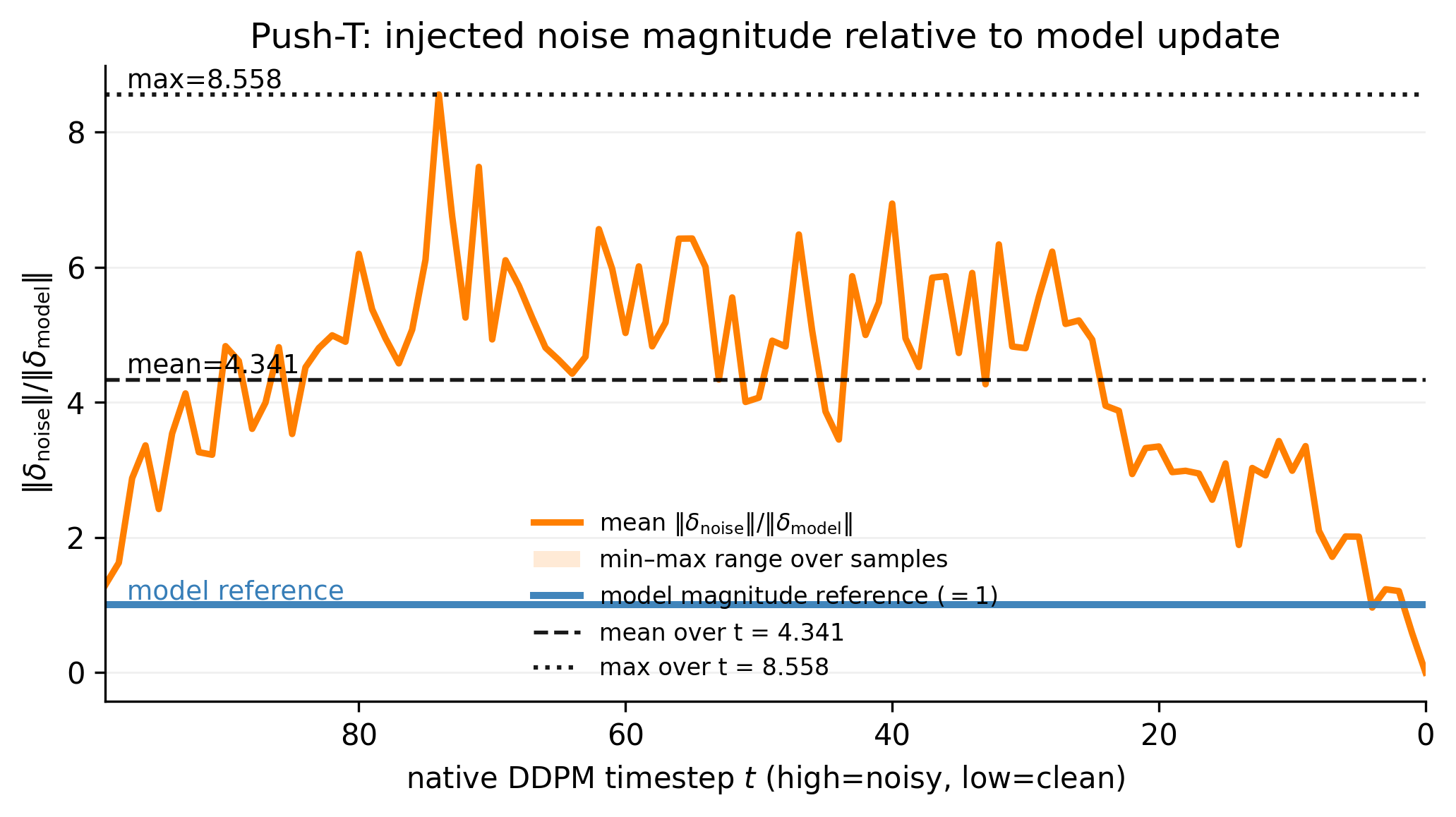}

        \vspace{0.3em}
        \textbf{(b) PushT}
    \end{minipage}

    \caption{
        Ratio between the magnitude of the stochastic noise term and the model
        update across denoising for Maze2D and PushT. These measurements were
        used as a reference for choosing an initial range of amplification
        factors, rather than to determine $\lambda$ directly.
    }
    \label{fig:model_noise_ratio}
\end{figure*}

\FloatBarrier
\section{Stochastic Amplification - Full Tabular Results}
\label{app:full_stoch_amp_results}

This appendix provides a comprehensive set of stochastic-amplification results compared to the sub-setted Table ~\ref{tab:stochastic_scaling}. The tables show how feasibility, diversity, and trajectory quality change as the noise scale $\lambda$ and intervention length $K$ are varied.

Maze2D is reported for a single example task, while MPD and PushT are reported across held-out task instances and therefore include mean $\pm$ SD. For MPD and PushT, undefined diversity or cost values are handled according to the metric definitions in Section~\ref{para:metrics}. These tables show the broader pattern of how stronger or longer amplification affects the balance between diversity, feasibility, and solution quality.

\begin{table}[ht]

\caption{
Maze2D stochastic-amplification results performed on a single held-out task. \(F\) is the number of feasible samples out of 50, \(D\) is feasible-sample diversity, and \(J\) is mean feasible-sample cost. The shared DDPM baseline is \(\lambda=1\); values below and above one respectively attenuate and amplify sampler noise.
}

\label{tab:maze2d_amplification_full}

\begin{center}

\small
\setlength{\tabcolsep}{3pt}

\begin{tabular*}{\textwidth}{
    @{\extracolsep{\fill}}
    c
    rrr
    rrr
    rrr
    rrr
}
\toprule
&
\multicolumn{3}{c}{$K=5,\;t=242{:}238$}
&
\multicolumn{3}{c}{$K=10,\;t=242{:}233$}
&
\multicolumn{3}{c}{$K=20,\;t=245{:}226$}
&
\multicolumn{3}{c}{$K=20,\;t=252{:}233$}
\\
\cmidrule(lr){2-4}
\cmidrule(lr){5-7}
\cmidrule(lr){8-10}
\cmidrule(lr){11-13}

$\lambda$
& $F\uparrow$ & $D\uparrow$ & $J\downarrow$
& $F\uparrow$ & $D\uparrow$ & $J\downarrow$
& $F\uparrow$ & $D\uparrow$ & $J\downarrow$
& $F\uparrow$ & $D\uparrow$ & $J\downarrow$
\\
\midrule

0.50
& 47 & 31.5 & 34.22
& 44 & 32.0 & 33.95
& 48 & 29.5 & 34.33
& 47 & 28.8 & 34.53
\\

0.75
& 46 & 33.2 & 34.22
& 45 & 33.0 & 34.25
& 46 & 30.0 & 34.16
& 48 & 30.5 & 34.34
\\

\midrule

\textbf{1.00}
& \textbf{44} & \textbf{35.2} & \textbf{34.19}
& \textbf{44} & \textbf{35.2} & \textbf{34.19}
& \textbf{44} & \textbf{35.2} & \textbf{34.19}
& \textbf{44} & \textbf{35.2} & \textbf{34.19}
\\

\midrule

1.50
& 45 & 37.0 & 34.15
& 38 & 39.4 & 33.24
& 38 & 43.8 & 33.34
& 39 & 46.8 & 33.96
\\

2.00
& 40 & 40.8 & 33.85
& 37 & 44.9 & 32.91
& 35 & 47.7 & 32.91
& 33 & 47.8 & 33.55
\\

2.50
& 37 & 44.3 & 33.45
& 34 & 45.6 & 33.02
& 31 & 47.5 & 33.76
& 33 & 50.0 & 32.66
\\

3.00
& 35 & 47.7 & 33.32
& 33 & 50.0 & 32.64
& 27 & 48.9 & 33.25
& 35 & 49.2 & 33.23
\\

5.00
& 28 & 49.8 & 33.43
& 25 & 44.9 & 33.56
& 23 & 44.2 & 33.41
& 20 & 47.9 & 30.96
\\

7.00
& 24 & 42.0 & 34.83
& 18 & 43.9 & 32.23
& 24 & 43.7 & 33.06
& 15 & 41.3 & 32.42
\\

10.00
& 26 & 43.7 & 34.01
& 17 & 39.4 & 33.23
& 20 & 38.8 & 32.97
& 16 & 37.6 & 33.34
\\

\bottomrule
\end{tabular*}

\end{center}

\end{table}

\begin{table}[ht]

\caption{
Expanded stochastic-amplification results for MPD-P and MPD-G.
Values are mean \(\pm\) SD across six held-out tasks, with repeated
run seeds averaged within each task before cross-task aggregation.
\(F\) is the number of feasible samples out of 50, \(D\) is feasible-sample
diversity, \(J\) is mean feasible-sample cost, and \(J^*\) is best
feasible-sample cost. Diversity is undefined when fewer than two
feasible samples are available, and cost is undefined when no feasible
samples are available.
}

\label{tab:mpd_amplification_expanded}

\begin{center}

\small
\setlength{\tabcolsep}{5pt}
\renewcommand{\arraystretch}{0.98}

\begin{tabular*}{\textwidth}{
    @{\extracolsep{\fill}}
    l
    rrrr
}
\toprule

& $F\uparrow$
& $D\uparrow$
& $J\downarrow$
& $J^*\downarrow$
\\
\midrule

\multicolumn{5}{l}{\textbf{MPD-P}} \\
\midrule

Baseline ($\lambda=1$)
& $19.40 \pm 21.16$
& $4.75 \pm 1.50$
& $12.76 \pm 4.38$
& $11.22 \pm 4.49$
\\

$(K=3,\ \lambda=7)$
& $17.59 \pm 14.17$
& $10.55 \pm 0.99$
& $14.74 \pm 2.76$
& $11.31 \pm 2.41$
\\

$(K=5,\ \lambda=4)$
& $20.20 \pm 15.77$
& $9.89 \pm 0.82$
& $14.12 \pm 3.03$
& $10.91 \pm 3.58$
\\

$(K=5,\ \lambda=7)$
& $16.04 \pm 11.63$
& $11.23 \pm 0.90$
& $15.17 \pm 2.65$
& $11.51 \pm 1.87$
\\

$(K=5,\ \lambda=10)$
& $13.40 \pm 10.82$
& $11.94 \pm 0.90$
& $16.71 \pm 2.69$
& $12.80 \pm 1.91$
\\

$(K=10,\ \lambda=4)$
& $19.46 \pm 14.05$
& $10.82 \pm 1.00$
& $14.30 \pm 2.73$
& $10.75 \pm 2.73$
\\

$(K=10,\ \lambda=5)$
& $16.38 \pm 11.16$
& $11.98 \pm 0.89$
& $15.25 \pm 2.46$
& $11.43 \pm 1.70$
\\

$(K=10,\ \lambda=7)$
& $6.71 \pm 6.20$
& $13.94 \pm 0.68$
& $22.58 \pm 2.89$
& $17.39 \pm 2.47$
\\

\midrule

\multicolumn{5}{l}{\textbf{MPD-G}} \\
\midrule

Baseline ($\lambda=1$)
& $47.77 \pm 5.12$
& $6.62 \pm 3.06$
& $13.98 \pm 3.78$
& $9.80 \pm 2.03$
\\

$(K=3,\ \lambda=3)$
& $47.27 \pm 5.73$
& $9.60 \pm 2.72$
& $14.17 \pm 2.86$
& $9.23 \pm 1.59$
\\

$(K=3,\ \lambda=7)$
& $45.94 \pm 5.51$
& $11.23 \pm 1.83$
& $15.35 \pm 2.14$
& $10.13 \pm 1.20$
\\

$(K=5,\ \lambda=4)$
& $46.72 \pm 5.71$
& $10.92 \pm 2.11$
& $14.70 \pm 2.15$
& $9.41 \pm 1.31$
\\

$(K=10,\ \lambda=4)$
& $45.75 \pm 6.18$
& $11.57 \pm 1.88$
& $14.94 \pm 2.00$
& $9.59 \pm 1.27$
\\

$(K=10,\ \lambda=5)$
& $41.72 \pm 7.26$
& $12.65 \pm 1.50$
& $15.83 \pm 1.90$
& $10.32 \pm 1.09$
\\

$(K=10,\ \lambda=7)$
& $16.70 \pm 7.68$
& $14.45 \pm 0.89$
& $23.91 \pm 2.39$
& $15.29 \pm 1.14$
\\

\bottomrule
\end{tabular*}

\end{center}

\end{table}

\begin{table}[!t]

\caption{
Expanded stochastic-amplification results for the PushT Transformer
and CNN policies. Values are mean \(\pm\) SD across 10 held-out task
instances for the Transformer and five for the CNN.
\(F\) is the number of successful closed-loop rollouts per task
(out of 56 for the Transformer and 50 for the CNN), \(D\) is block-pose
diversity over successful rollouts, and \(J\) is the mean number of
environment steps to success. For each \(K\), amplification is applied
over the final \(K\) denoising transitions. \(D\) is undefined for a task
with fewer than two successful rollouts, and \(J\) is undefined when no
successful rollout is available. Aggregate statistics use the task
instances for which the corresponding metric is defined; where only
one task defines a metric, only the mean is reported.
}

\label{tab:pusht_amplification_expanded}

\begin{center}

\small
\setlength{\tabcolsep}{5pt}
\renewcommand{\arraystretch}{0.98}

\begin{tabular*}{\textwidth}{
    @{\extracolsep{\fill}}
    l
    rrr
}
\toprule

& $F\uparrow$
& $D\uparrow$
& $J\downarrow$
\\
\midrule

\multicolumn{4}{l}{\textbf{PushT Transformer}} \\
\midrule

Baseline ($\lambda=1$)
& $17.2 \pm 13.55$
& $352.03 \pm 211.41$
& $204.13 \pm 39.45$
\\

$(K=5,\ \lambda=1.5)$
& $16.3 \pm 11.95$
& $485.89 \pm 495.60$
& $201.55 \pm 47.03$
\\

$(K=5,\ \lambda=2)$
& $15.9 \pm 11.99$
& $500.05 \pm 353.73$
& $199.65 \pm 46.18$
\\

$(K=5,\ \lambda=3)$
& $13.6 \pm 10.63$
& $502.70 \pm 251.69$
& $203.15 \pm 41.74$
\\

$(K=5,\ \lambda=5)$
& $2.7 \pm 2.11$
& $647.14 \pm 213.85$
& $199.61 \pm 51.82$
\\

$(K=5,\ \lambda=7)$
& $0.3 \pm 0.48$
& --
& $171.33 \pm 101.01$
\\

$(K=5,\ \lambda=10)$
& $0.0 \pm 0.00$
& --
& --
\\

$(K=5,\ \lambda=25)$
& $0.0 \pm 0.00$
& --
& --
\\

$(K=10,\ \lambda=1.5)$
& $17.8 \pm 15.22$
& $414.42 \pm 432.93$
& $198.66 \pm 55.48$
\\

$(K=10,\ \lambda=2)$
& $16.2 \pm 12.39$
& $546.51 \pm 403.06$
& $191.65 \pm 39.61$
\\

$(K=10,\ \lambda=3)$
& $10.7 \pm 8.19$
& $632.85 \pm 264.68$
& $204.66 \pm 37.94$
\\

$(K=10,\ \lambda=5)$
& $0.6 \pm 0.70$
& $1142.68$
& $209.70 \pm 74.07$
\\

$(K=10,\ \lambda=7)$
& $0.0 \pm 0.00$
& --
& --
\\

$(K=10,\ \lambda=10)$
& $0.0 \pm 0.00$
& --
& --
\\

$(K=10,\ \lambda=25)$
& $0.0 \pm 0.00$
& --
& --
\\

$(K=20,\ \lambda=1.5)$
& $16.9 \pm 12.36$
& $390.30 \pm 234.11$
& $201.29 \pm 42.10$
\\

$(K=20,\ \lambda=2)$
& $15.3 \pm 11.91$
& $528.68 \pm 442.62$
& $196.65 \pm 49.86$
\\

$(K=20,\ \lambda=3)$
& $6.6 \pm 5.48$
& $637.65 \pm 184.57$
& $196.07 \pm 31.79$
\\

$(K=20,\ \lambda=5)$
& $0.2 \pm 0.63$
& $625.02$
& $159.00$
\\

$(K=20,\ \lambda=7)$
& $0.0 \pm 0.00$
& --
& --
\\

$(K=20,\ \lambda=10)$
& $0.0 \pm 0.00$
& --
& --
\\

$(K=20,\ \lambda=25)$
& $0.0 \pm 0.00$
& --
& --
\\

\midrule

\multicolumn{4}{l}{\textbf{PushT CNN}} \\
\midrule

Baseline ($\lambda=1$)
& $18.8 \pm 22.35$
& $425.34 \pm 530.15$
& $201.81 \pm 85.75$
\\

$(K=5,\ \lambda=1.5)$
& $17.8 \pm 20.97$
& $338.69 \pm 317.77$
& $198.77 \pm 67.24$
\\

$(K=5,\ \lambda=2)$
& $17.0 \pm 19.84$
& $465.19 \pm 483.38$
& $197.42 \pm 51.66$
\\

$(K=5,\ \lambda=3)$
& $14.6 \pm 15.96$
& $583.34 \pm 632.60$
& $183.63 \pm 39.18$
\\

$(K=5,\ \lambda=5)$
& $3.0 \pm 2.65$
& $856.38 \pm 230.36$
& $212.43 \pm 51.05$
\\

$(K=10,\ \lambda=1.5)$
& $18.6 \pm 22.56$
& $418.04 \pm 448.76$
& $174.20 \pm 64.79$
\\

$(K=10,\ \lambda=2)$
& $16.8 \pm 20.90$
& $357.37 \pm 159.22$
& $209.15 \pm 70.60$
\\

$(K=10,\ \lambda=3)$
& $14.8 \pm 17.12$
& $709.66 \pm 493.50$
& $228.30 \pm 66.38$
\\

$(K=10,\ \lambda=5)$
& $0.6 \pm 0.89$
& $321.67$
& $223.75 \pm 51.27$
\\

$(K=20,\ \lambda=1.5)$
& $16.6 \pm 21.48$
& $406.15 \pm 418.02$
& $192.41 \pm 58.70$
\\

$(K=20,\ \lambda=2)$
& $17.8 \pm 19.75$
& $498.18 \pm 544.53$
& $190.21 \pm 56.36$
\\

$(K=20,\ \lambda=3)$
& $10.2 \pm 11.32$
& $834.81 \pm 657.06$
& $210.96 \pm 53.25$
\\

$(K=20,\ \lambda=5)$
& $0.0 \pm 0.00$
& --
& --
\\

\bottomrule
\end{tabular*}

\end{center}

\end{table}

\FloatBarrier
\section{Stochastic Amplification - Qualitative Examples}
\label{app:stoch_amp_qualitative}

We provide further qualitative examples of stochastic amplification across the three tested benchmarks. 
As intervention strength increases, a point is reached where the increase in diversity is not accompanied with an increase of any of the other performance metrics. We consider a reduction in batch level feasibility or sample quality an acceptable trade-off only if it is also accompanied with either emergence of a new sample solution, or an improvement in one other metric, such as best-of-\(N\) sample cost.


\begin{figure*}[h]
    \centering
    \includegraphics[
        width=\textwidth
    ]{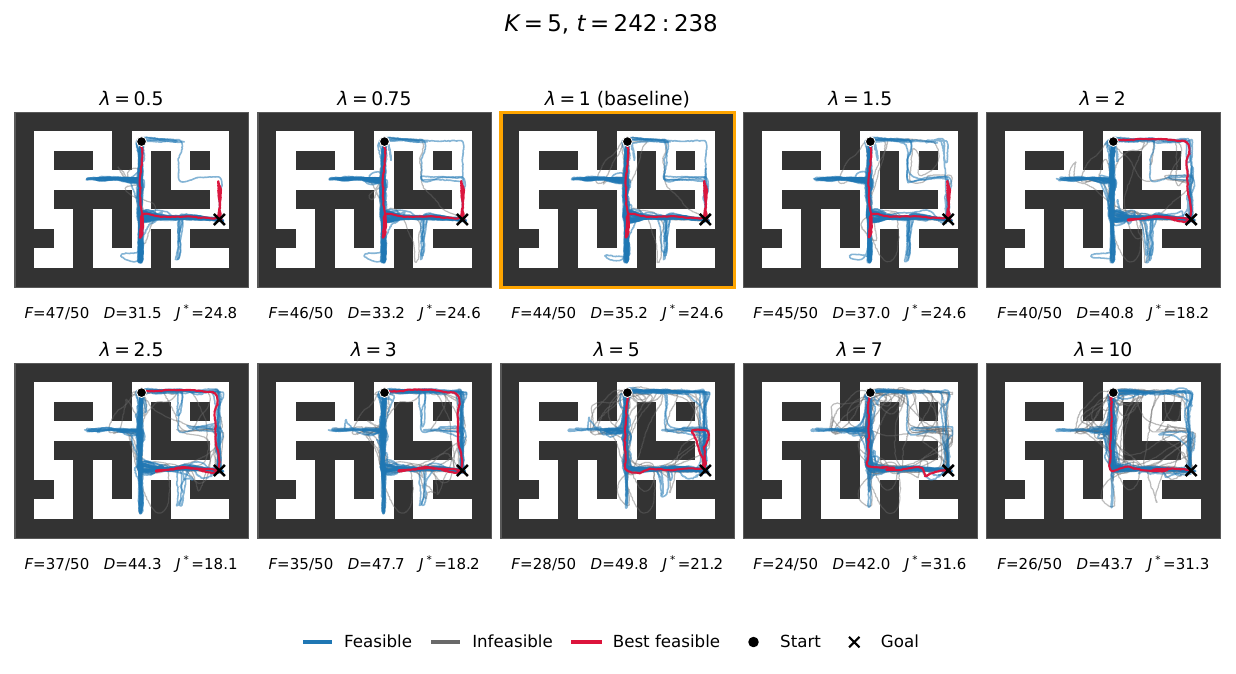}
    \caption{Qualitative Maze2D samples under stochastic amplification for the
    budgeted target interval \(K=5\), \(t=242{:}238\). Increasing the scale factor results in greater exploration of the solution space across feasible samples the batch, the trade-off that is evident is that although many of those samples will be longer paths, some also exhibit shorter, more efficient paths than the baseline, thus batch sampling with extra stochastic pressure provides opportunity to yield improved best samples.}
    \label{fig:maze2d_amplification_qualitative_k5}
\end{figure*}

[\begin{figure*}[!tbp]]
    \centering
    \includegraphics[
        width=\textwidth
    ]{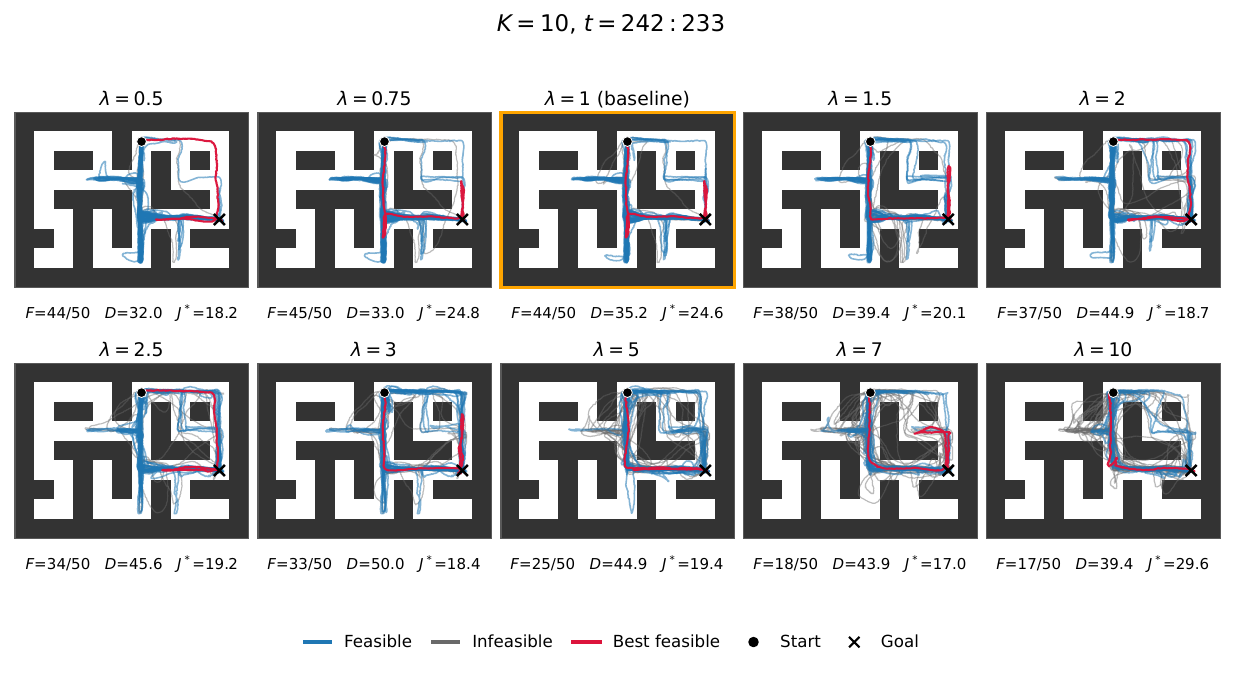}
    \caption{Qualitative Maze2D samples under stochastic amplification for the
    budgeted target interval \(K=10\), \(t=242{:}233\). The trade-offs from increased stochastic pressure are even clearer. Our targeted diagnostic intervals provide a parsimonious target regime to apply scale factoring.}
    \label{fig:maze2d_amplification_qualitative_k10}
\end{figure*}

\begin{figure*}[!tbp]
    \centering
    \includegraphics[
        width=\textwidth
    ]{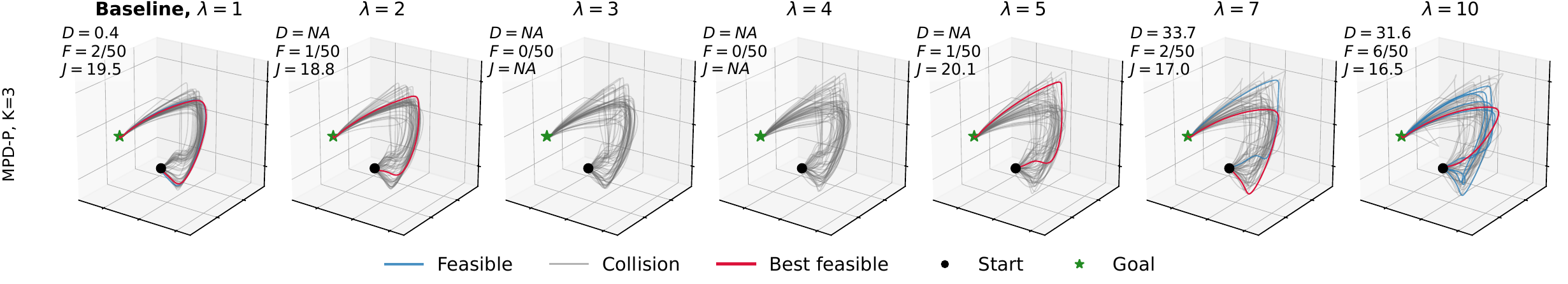}

    \smallskip
    \textbf{(a) MPD-P}

    \medskip

    \includegraphics[
        width=\textwidth
    ]{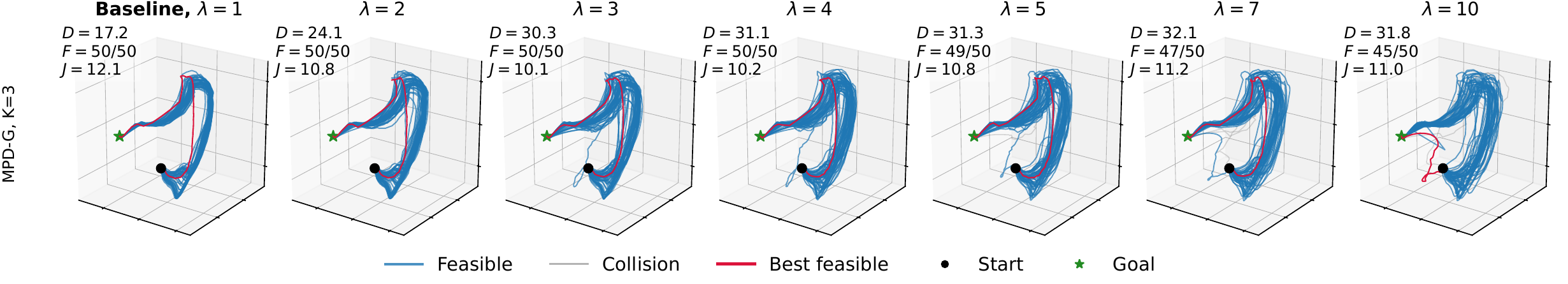}

    \smallskip
    \textbf{(b) MPD-G}

    \caption{Qualitative MPD stochastic-amplification results for the budgeted target interval \(K=3\), \(t=24{:}22\). The rows compare the unguided prior, MPD-P, and cost-guided sampling, MPD-G, across the baseline and selected noise scale factors \(\lambda\). \(F\), \(D\), and \(J\) denote the feasible count, feasible-trajectory diversity, and best feasible-trajectory cost, respectively. At \(\lambda\) = 10 (for K=3),for MPD-G, a new route solution is found that out-performs all others of its batch.}
    \label{fig:mpd_amplification_qualitative_k3}
\end{figure*}

\begin{figure*}[!tbp]
    \centering
    \includegraphics[
        width=\textwidth
    ]{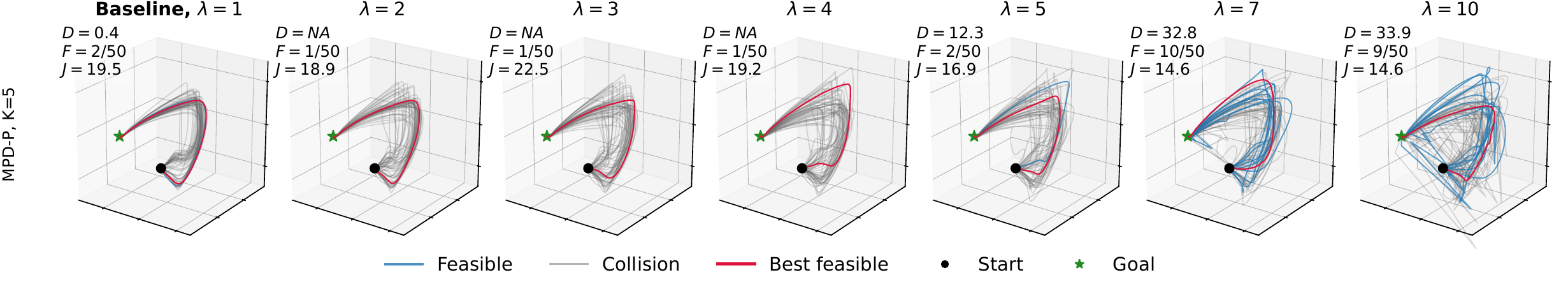}

    \smallskip
    \textbf{(a) MPD-P}

    \medskip

    \includegraphics[
        width=\textwidth
    ]{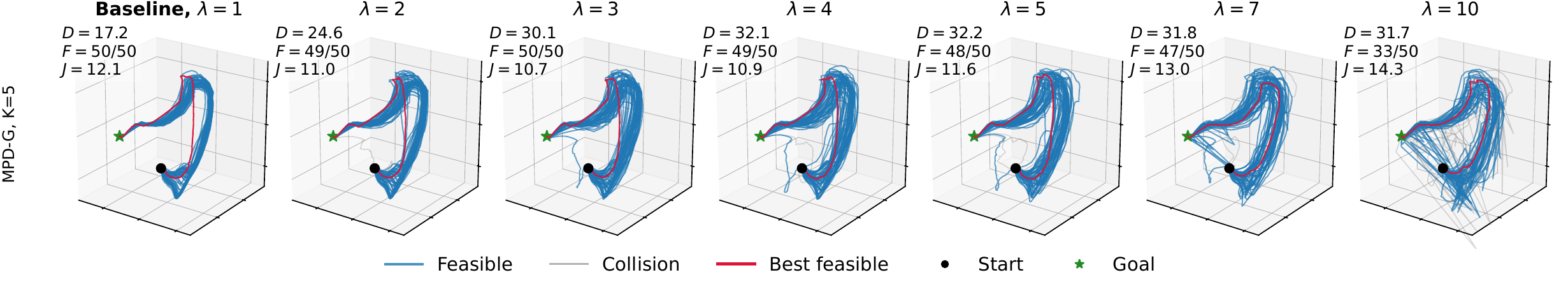}

    \smallskip
    \textbf{(b) MPD-G}

    \caption{Qualitative MPD stochastic-amplification results for the budgeted target interval \(K=5\), \(t=24{:}20\). The rows compare MPD-P and MPD-G across the baseline and selected noise scale factors \(\lambda\). \(F\), \(D\), and \(J\) denote the feasible count, feasible-trajectory diversity, and best feasible-trajectory cost, respectively. At \(\lambda\) = 3 a new route solution is found for MPD-G}
    \label{fig:mpd_amplification_qualitative_k5}
\end{figure*}

\begin{figure*}[!tbp]
    \centering
    \includegraphics[
        width=\textwidth
    ]{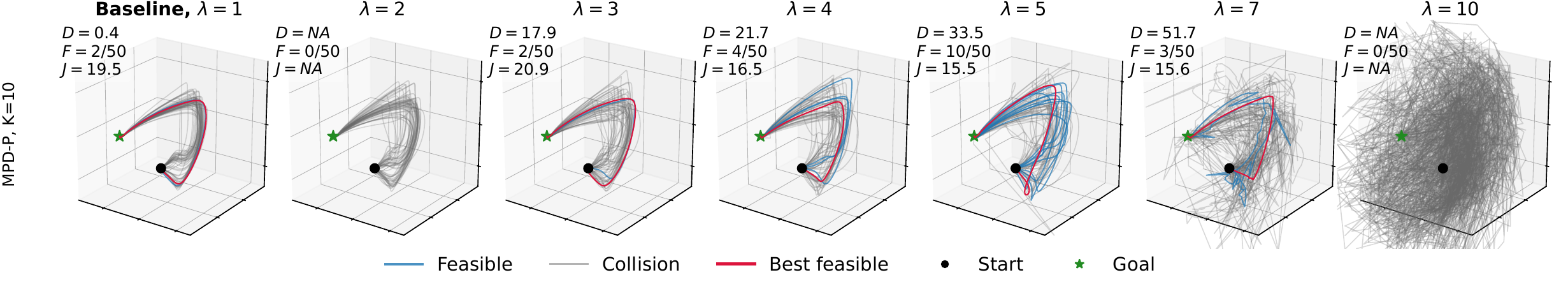}

    \smallskip
    \textbf{(a) MPD-P}

    \medskip

    \includegraphics[
        width=\textwidth
    ]{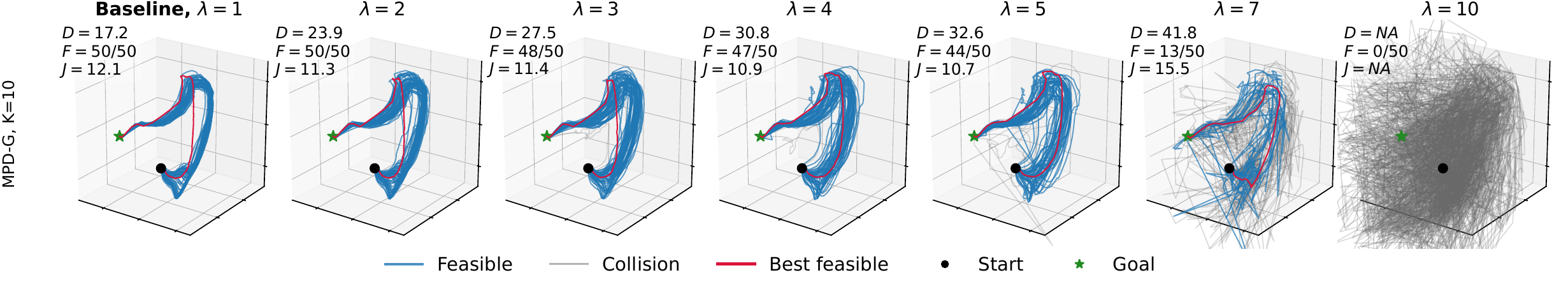}

    \smallskip
    \textbf{(b) MPD-G}

    \caption{Qualitative MPD stochastic-amplification results for the budgeted target interval \(K=10\), \(t=24{:}15\). The rows compare MPD-P and MPD-G across the baseline and selected noise scale factors \(\lambda\). \(F\), \(D\), and \(J\) denote the feasible count, feasible-trajectory diversity, and best feasible-trajectory cost, respectively. We can see that too much stochastic pressure moves the sample out of supported region for the model to predict according to its learned objectives.}
    \label{fig:mpd_amplification_qualitative_k10}
\end{figure*}

\begin{figure*}[!tbp]
    \centering

    \includegraphics[
        width=0.9\textwidth
    ]{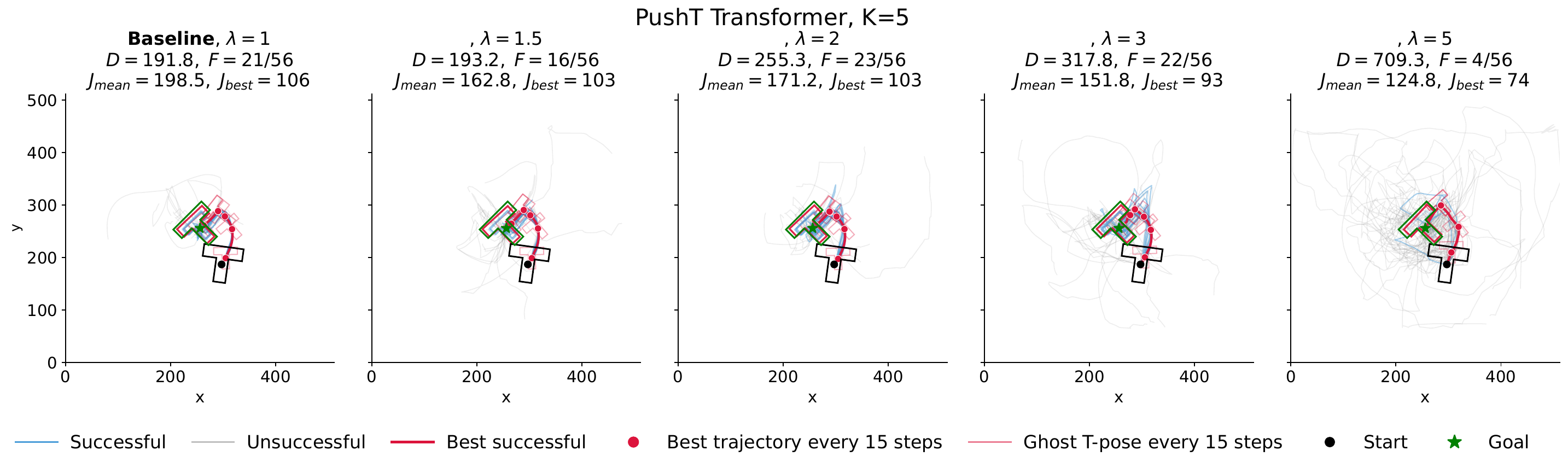}

    \vspace{0.1em}
    \textbf{(a) PushT Transformer}

    \vspace{0.3em}

    \includegraphics[
        width=0.9\textwidth
    ]{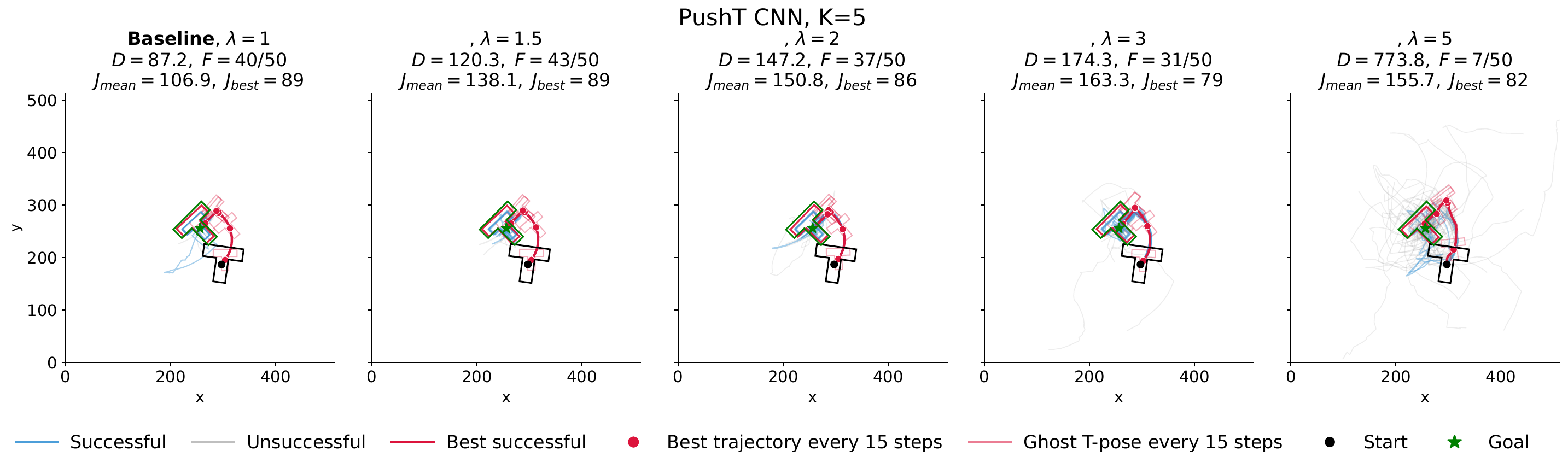}

    \vspace{0.1em}
    \textbf{(b) PushT CNN}

    \caption{Qualitative PushT stochastic-amplification results for the budgeted target interval \(K=5\), \(t=5{:}1\), shown for episode 4. The rows compare the Transformer and CNN policies across the baseline and selected noise scale factors \(\lambda\). The improvement to sample quality (cost) is largely seen in fewer, but stronger pushes along a similar route.}
    \label{fig:pusht_amplification_qualitative_k5}
\end{figure*}

\begin{figure*}[!tbp]
    \centering
    \includegraphics[
        width=0.9\textwidth
    ]{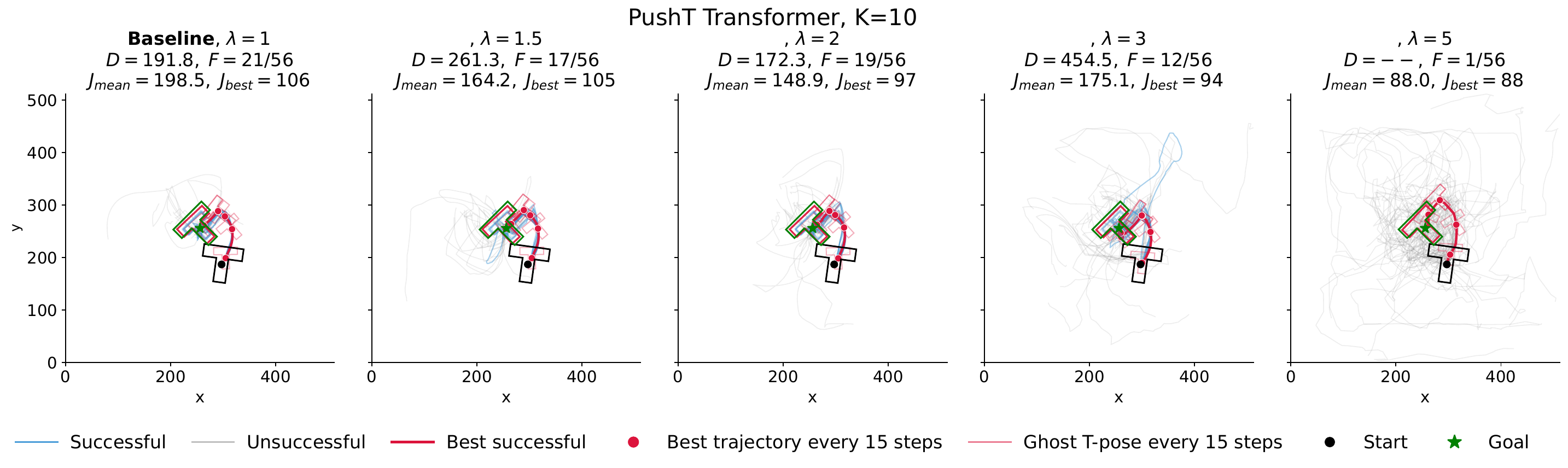}

    \smallskip
    \textbf{(a) PushT Transformer}

    \medskip

    \includegraphics[
        width=0.9\textwidth
    ]{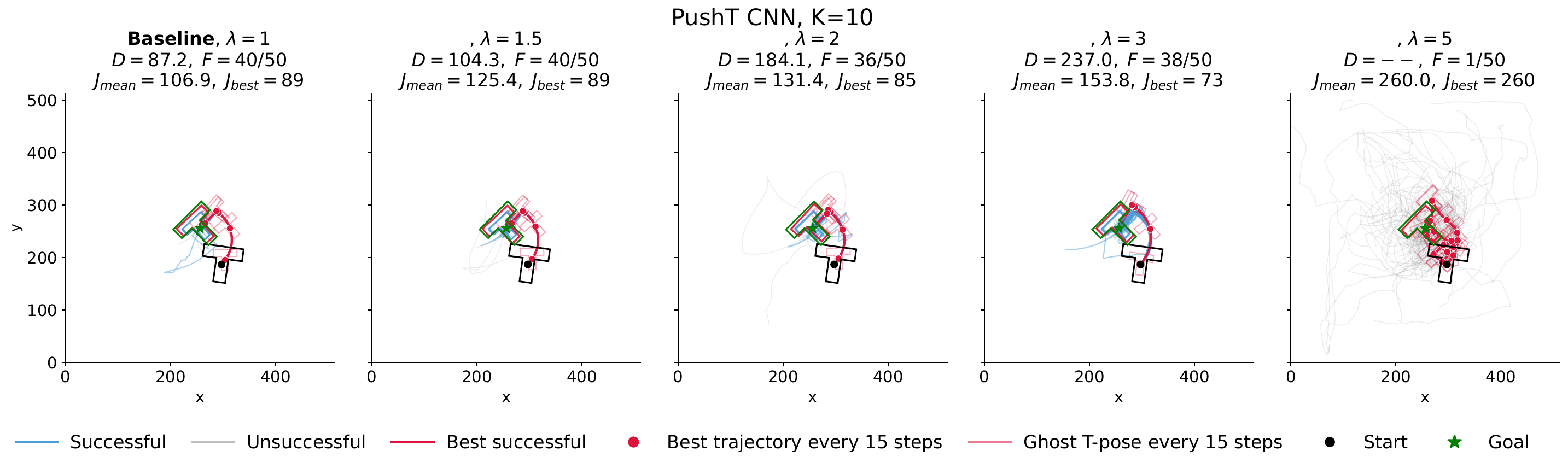}

    \smallskip
    \textbf{(b) PushT CNN}

    \caption{Qualitative PushT stochastic-amplification results for the budgeted target interval \(K=10\), \(t=10{:}1\), shown for episode 4.}
    \label{fig:pusht_amplification_qualitative_k10}
\end{figure*}

\begin{figure*}[!tbp]
    \centering
    \includegraphics[
        width=0.85\textwidth
    ]{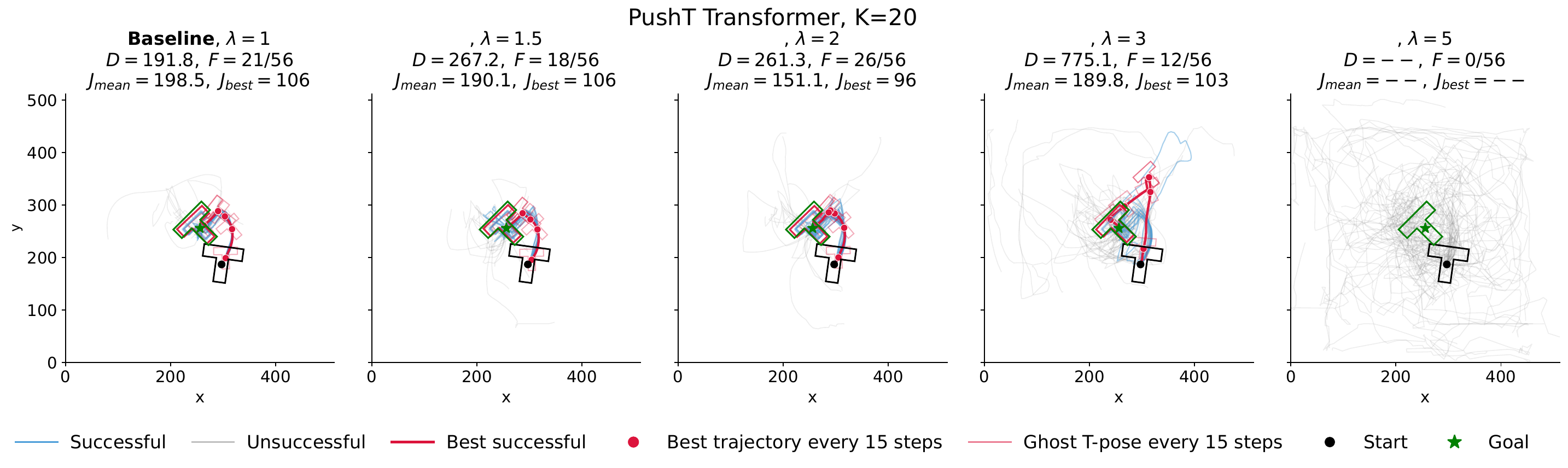}

    \smallskip
    \textbf{(a) PushT Transformer}

    \medskip

    \includegraphics[
        width=0.85\textwidth
    ]{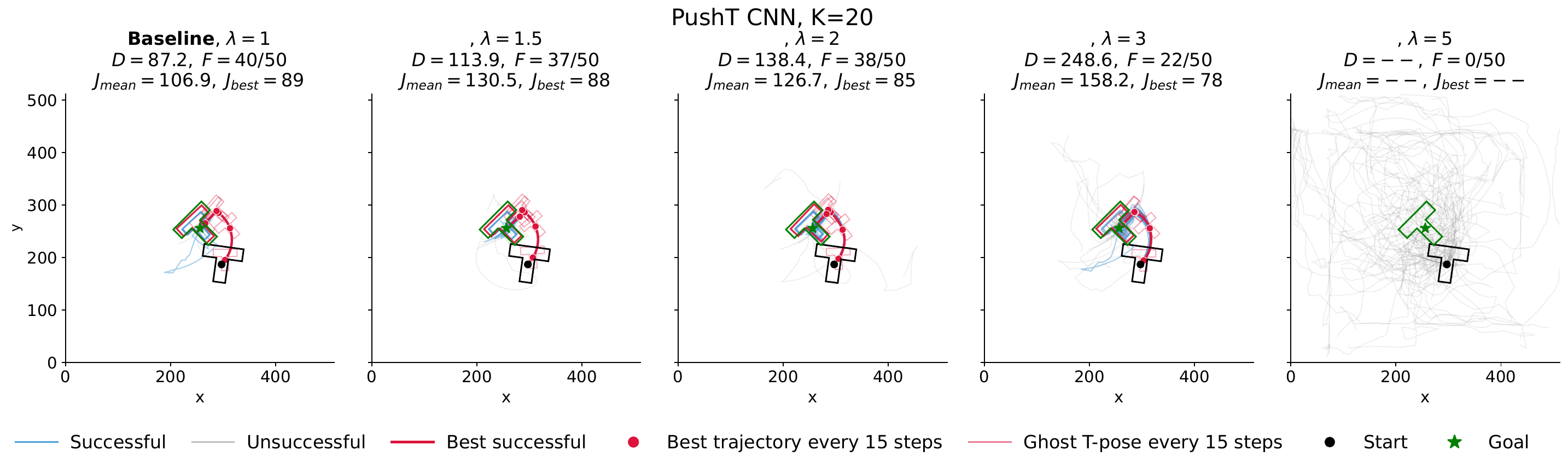}

    \smallskip
    \textbf{(b) PushT CNN}

    \caption{Qualitative PushT stochastic-amplification results for the budgeted target interval \(K=20\), \(t=20{:}1\), shown for episode 4.}
    \label{fig:pusht_amplification_qualitative_k20}
\end{figure*}

\FloatBarrier
\section{SDW-Hybrid Timesteps}
\label{app:SDWHybrid_steps}

Table~\ref{tab:hybrid_schedules} reports the DDPM timesteps
retained by the SDW-hybrid sampler and the matched middle-window controls of Section~\ref{sec: SDW-Hybrid}. All other regions are traversed using deterministic DDIM skips with a nominal stride of five shortened where necessary to preserve retained DDPM transitions. The MPD-G configuration additionally retains the late cost-guidance region in both schedules.

\begin{table}[ht]
\centering
\caption{Retained DDPM timesteps for SDW-hybrid and
matched middle-window controls. Timesteps are listed
in reverse denoising order (noisy to clean). NFEs are
identical between the corresponding schedules.}
\label{tab:hybrid_schedules}
\small
\begin{tabular}{llll}
\toprule
Configuration & SDW-hybrid & Middle control & NFEs \\
\midrule
MPD-P, $K=5$  & $24$--$20$ & $14$--$10$ & 10 \\
MPD-P, $K=10$ & $24$--$15$ & $14$--$5$  & 14 \\
MPD-G, $K=5$  & $24$--$20$, $6$--$0$
              & $14$--$10$, $6$--$0$ & 15 \\
\bottomrule
\end{tabular}
\end{table}

\end{document}

%% file: math_commands.tex
\usepackage{amsmath,amsfonts,bm}

\def\eqref#1{equation~\ref{#1}}

\def\1{\bm{1}}

\DeclareMathAlphabet{\mathsfit}{\encodingdefault}{\sfdefault}{m}{sl}
\SetMathAlphabet{\mathsfit}{bold}{\encodingdefault}{\sfdefault}{bx}{n}

